\pdfoutput=1
\documentclass{article}

\usepackage{microtype}
\usepackage{graphicx}
\graphicspath{{figures/}}
\usepackage{booktabs}
\usepackage{array}
\usepackage{amssymb}
\usepackage{amsmath}
\usepackage{xcolor}
\usepackage{framed}
\usepackage{multirow}
\usepackage{algorithm}
\usepackage{algorithmic}
\usepackage{url}
\usepackage{placeins}
\usepackage{enumitem}
\usepackage{listings}
\DeclareMathOperator*{\argmax}{arg\,max}

\usepackage{hyperref}

\usepackage[accepted]{icml2024}

\newcommand{\ygold}{y^{\mathrm{gold}}}
\newcommand{\pTS}{T \succ S}
\newcommand{\pST}{S \succ T}
\newcommand{\pSS}{S \succ S}
\newcommand{\mc}[2][c]{\begin{tabular}[c]{@{}#1@{}}#2\end{tabular}}

\icmltitlerunning{An Empirical Study of Direct Nash Optimization}

\begin{document}

\twocolumn[
\icmltitle{Aligning Language Models with General Preferences: \\ An Empirical Study of Direct Nash Optimization}

\begin{icmlauthorlist}
\icmlauthor{Austin Tsai}{snps}
\icmlauthor{Saurav Nanda}{snps}
\icmlauthor{Ganapathy Parthasarathy}{snps}
\icmlauthor{Bhuvnesh Kumar}{snps}
\icmlauthor{Parivesh Choudhary}{snps}
\end{icmlauthorlist}

\icmlaffiliation{snps}{Synopsys, Inc., Sunnyvale, CA, USA}

\icmlcorrespondingauthor{Bhuvnesh Kumar}{bhuvnesh@synopsys.com}

\icmlkeywords{RLHF, preference optimization, Direct Nash Optimization, DPO, LLM alignment, AlpacaEval, LLM-as-a-judge}

\vskip 0.3in
]

\hypersetup{
  pdftitle={Aligning Language Models with General Preferences: An Empirical Study of Direct Nash Optimization},
  pdfauthor={Austin Tsai, Saurav Nanda, Ganapathy Parthasarathy, Bhuvnesh Kumar, Parivesh Choudhary},
  pdfsubject={arXiv preprint},
  pdfkeywords={RLHF, preference optimization, Direct Nash Optimization, DPO, LLM alignment, AlpacaEval, LLM-as-a-judge}}

\printAffiliationsAndNotice{Work done while all authors were at Synopsys, Inc.\ }

\begin{abstract}
Direct Nash Optimization (DNO) aligns a language model to a general preference function by approximating the Nash equilibrium of a two-player preference game. Its practical variant, DNO-Prct, iterates on-policy sampling, LLM-based rating of the samples together with a teacher's gold response, large-margin preference-pair selection and a DPO-style contrastive update. We re-implement DNO-Prct independently and study it on two open-weight instruct models, Llama~3.1 8B Instruct and Mistral~7B Instruct~v0.1, with GPT-4o or a larger open-weight model of the same family as the annotator. On AlpacaEval~2.0 with GPT-4o as judge, three iterations raise the length-controlled (LC) win rate against GPT-4o from 31.7\% to 45.3\% for Llama and from 6.8\% to 18.4\% for Mistral, and the tuned 8B model scores above Llama~3.1 70B Instruct on every reported win rate. With same-family open-weight annotators, the runs retain 62\% (Llama) and 82\% (Mistral) of the LC gain achieved with GPT-4o as annotator, so the rating step does not require a proprietary model. In a separate ablation configuration, variants that remove the gold responses, remove the data look-back, or accept only teacher-preferred pairs score 6 to 10 LC points below the full procedure, whereas varying the number of on-policy samples per prompt (1 to 7) moves the LC win rate by at most 2.1 points, and a margin sweep (1 to 5) run in its own pipeline by at most 2.4 points. We also report the cost of the pipeline and where the practical algorithm departs from the theory it approximates. All results are single runs, and GPT-4o is both the teacher and the evaluation judge; we discuss what this implies for the reported gains.
\end{abstract}

\section{Introduction}
\label{sec:intro}

Pre-trained large language models (LLMs) have advanced rapidly \citep{radford2018gpt,radford2019gpt2,brown2020gpt3,rae2021gopher}. Trained only to predict the next token, they are not by themselves aligned with what humans prefer. Reinforcement Learning from Human Feedback (RLHF) \citep{christiano2017deep,stiennon2020summarize,ouyang2022instructgpt,bai2022hh,touvron2023llama2,openai2023gpt4} bridges this gap by collecting human preferences, training a reward model on them, and optimizing the policy with reinforcement learning. Single-stage contrastive methods such as Direct Preference Optimization \citep[DPO;][]{rafailov2023dpo} remove the explicit reward model. In their standard formulations, both assume a pointwise (Bradley--Terry) reward and therefore cannot represent intransitive preferences, which \citet{liu2025condorcet} show arise with high probability under a Luce-type preference model; they are also vulnerable to reward hacking \citep{amodei2016concrete}, and a reward model or preference dataset fixed in advance becomes stale as the policy distribution shifts during training \citep{ross2011dagger,cheng2024apo,azar2023ipo,munos2023nlhf}.

To address these limitations, a line of work optimizes the general preference function directly by finding the Nash equilibrium of a two-player game, solving that game with single-player (self-play) algorithms that exploit the symmetry of the preference function \citep{munos2023nlhf,swamy2024spo}. For a preference function with \mbox{$\mathcal{P}(y \succ y' \mid x) + \mathcal{P}(y' \succ y \mid x) = 1$} and $\mathcal{P}(\pi \succ \pi') := \mathbb{E}_{x \sim \rho,\, y \sim \pi(\cdot \mid x),\, y' \sim \pi'(\cdot \mid x)}[\mathcal{P}(y \succ y' \mid x)]$ for a prompt distribution $\rho$, a Nash equilibrium is a policy $\pi^{\star}$ with $\mathcal{P}(\pi^{\star} \succ \pi) \ge \tfrac12$ for every competing policy $\pi$ in the policy class $\Pi$; this policy is the Minimax Winner of the game. \citet{rosset2024dno} propose Direct Nash Optimization (DNO): a provable algorithm whose iterates converge to the Nash equilibrium on average, together with a practical variant, DNO-Prct, which in their largest run, on about ten times more prompts over six iterations, brought a 7B model to a 33\% length-controlled AlpacaEval~2.0 win rate against GPT-4 Turbo, with GPT-4 Turbo as annotator and gold-response source. DNO-Prct bundles several design choices: batched on-policy sampling, a teacher's gold response among the candidates, LLM-based rating against a rubric, a minimum rating margin, and a decaying look-back over earlier pairs.

To our knowledge no independent re-implementation of DNO-Prct has been published: its published results, including its ablations, use one base model and a proprietary annotator, so it is not known whether its gains and design choices transfer to other base models and annotators.
This paper is an empirical study of DNO-Prct. We re-implemented the algorithm from the description in \mbox{\citet{rosset2024dno}}, ran it on two open-weight instruct models with GPT-4o and with same-family open-weight models as annotators, and examined its main design choices. Throughout, ``DNO'' refers to DNO-Prct unless stated otherwise. Our contributions are as follows.

\textbf{(1) An independent re-implementation with a procedural description:} We give the pair-construction procedure as pseudo-code (Appendix~\ref{app:pairs}), the rating prompt verbatim (Appendix~\ref{app:prompt}) and the hyperparameters and implementation settings (Appendix~\ref{app:hparams}). Together with \citet{rosset2024dno}, this describes the pipeline in enough detail to re-implement it.

\textbf{(2) Gains on two model families:} Over three iterations the student's length-controlled (LC) AlpacaEval~2.0 win rate against GPT-4o rises from 31.71\% to 45.32\% for Llama~3.1 8B Instruct \citep{grattafiori2024llama3} and from 6.77\% to 18.44\% for Mistral~7B Instruct~v0.1 \citep{jiang2023mistral7b}. Under the same protocol, the tuned 8B model scores above Llama~3.1 70B Instruct, which has nearly nine times as many parameters, on all four reported win rates, and the tuned 7B model narrows its gap to Mixtral~8x7B Instruct~v0.1 \citep{jiang2024mixtral} on all four, to between 0.2 and 12.8 points (Section~\ref{sec:results:larger}).

\textbf{(3) Open-weight annotators:} Replacing GPT-4o with Llama~3.1 70B Instruct or Mistral Large~2 as the annotator retains 62\% (Llama) and 82\% (Mistral) of the LC gain against the GPT-4o opponent; against GPT-4-0613 the Llama run retains 45\% and the Mistral run matches the GPT-4o-annotated run, so the rating step does not require a proprietary model (Section~\ref{result:larger-annotators}).

\textbf{(4) Which components matter:} In a separate ablation configuration, variants that remove the gold responses, remove the data look-back, or restrict pair selection to teacher-preferred pairs score 6 to 10 LC points below the full procedure against GPT-4o (9 to 19 points against GPT-4-0613), whereas the number of on-policy samples per prompt (1 to 7) moves the LC win rate against GPT-4o by at most 2.1 points, and a margin sweep (1 to 5) run in its own pipeline by at most 2.4 points (Section~\ref{sec:ablation}). We also document the cost of the pipeline (Section~\ref{sec:costs}) and the places where DNO-Prct departs from the theory it approximates (Section~\ref{sec:ablation:errors}).

\citet[Section~5.2]{rosset2024dno} compared large-margin filtering, on-policy sampling and the SPIN and DNO-Restrictive pair rules in their own setting, and their Table~2 tracks teacher- and student-preferred pairs and the average teacher and student ratings across iterations. We extend these analyses beyond that setting, to two further base models and to open-weight annotators, add sweeps of the rating margin and the number of samples, and repeat their teacher--student analysis (Section~\ref{sec:teacher-student-framework}). The experiments were run in 2024, and Section~\ref{sec:related} relates them to work that appeared during or after them, notably on preference leakage between data generator and judge \citep{li2026preferenceleakage} and on the exploitability of AlpacaEval \citep{zheng2025nullmodels}.

\section{Preliminaries: Direct Nash Optimization}
\label{sec:dno}

There are two variants of Direct Nash Optimization: the theoretical DNO (Algorithm~\ref{alg:dno}) and the practical DNO-Prct. \citet{rosset2024dno} define the reward of a response at iteration $t$ as its expected win rate against the current policy's own samples under the preference function $\mathcal{P}$:
\begin{equation}
    r_t(x, y) := \mathbb{E}_{y' \sim \pi_t(\cdot \mid x)} \big[ \mathcal{P}(y \succ y' \mid x) \big],
\label{eq:reward-expectation}
\end{equation}
for prompts $x \in \mathcal{X}$ and responses $y \in \mathcal{Y}$. DNO aligns the implicit reward used by DPO with this preference-based reward. The implicit reward of a policy $\pi$ relative to $\pi_t$ is
\begin{align}
    r_{\pi,t}(x, y) &:= \eta \log \frac{\pi(y \mid x)}{\pi_t(y \mid x)} + \eta \log Z_t(x), \label{eq:internal-reward}\\
    Z_t(x) &:= \sum_{y \in \mathcal{Y}} \pi_t(y \mid x)\exp\!\Big(\frac{r_t(x,y)}{\eta}\Big), \nonumber
\end{align}
where $\eta > 0$ is the coefficient of the KL penalty in the objective $\mathcal{P}(\pi \succ \pi_t) - \eta\, \mathbb{E}_{x \sim \rho}[D_{\mathrm{KL}}(\pi \,\|\, \pi_t)]$, whose maximizer is the soft-policy-iteration update
\begin{equation*}
\pi^{\star}_{t+1}(y \mid x) = \pi_t(y \mid x)\exp\big(r_t(x,y)/\eta\big)/Z_t(x);
\end{equation*}
\citet{rosset2024dno} call $\eta$ the learning rate. In Eq.~\eqref{eq:dno-objective} below it plays the role of $\beta$ in DPO (in DNO-Prct, $\beta$ corresponds to a rescaled $\tilde{\eta}$; Section~\ref{sec:ablation:errors}), and it is distinct from the optimizer's learning rate. DNO minimizes a cross-entropy loss between the predicted preference $\sigma(r_{\pi,t}(x, y_1) - r_{\pi,t}(x, y_2))$ and the target preference $\sigma(r_t(x, y_1) - r_t(x, y_2))$, with $\sigma(z) := 1/(1+e^{-z})$. Writing
\begin{align}
\Delta_{\pi, t}(x, y_1, y_2) &:= r_{\pi,t}(x, y_1) - r_{\pi,t}(x, y_2) \nonumber\\
  &\;= \eta \log \tfrac{\pi(y_1 \mid x)}{\pi_t(y_1 \mid x)} - \eta \log \tfrac{\pi(y_2 \mid x)}{\pi_t(y_2 \mid x)}, \\
\Delta^{\star}_{t}(x, y_1, y_2) &:= r_t(x, y_1) - r_t(x, y_2),
\end{align}
in which the partition function cancels, the update at iteration $t$ is
\begin{equation}
\begin{aligned}
\pi_{t+1} = \argmax_{\pi \in \Pi}\; \mathbb{E}_{\mathcal{D}_t} \Big[\,
   & \sigma(\Delta^{\star}_{t}) \log \sigma(\Delta_{\pi, t}) \\
 + & \sigma(-\Delta^{\star}_{t}) \log \sigma(-\Delta_{\pi, t}) \Big],
\end{aligned}
\label{eq:dno-objective}
\end{equation}
where the arguments $(x, y_1, y_2)$ are omitted, the expectation is over $(x, y_1, y_2) \sim \mathcal{D}_t$, and the second term is the first with $y_1$ and $y_2$ exchanged.

\begin{algorithm}[t]
\caption{Direct Nash Optimization (adapted from \citealp{rosset2024dno})}
\label{alg:dno}
\begin{algorithmic}[1]
\REQUIRE preference function $\mathcal{P}$, KL coefficient $\eta$, iterations $N$, prompt distribution $\rho$
\STATE Initialize $\pi_1 \leftarrow \mathrm{unif}(\mathcal{Y})$.
\FOR{$t = 1, \ldots, N$}
  \STATE Compute $r_t(x, y)$ by Eq.~\eqref{eq:reward-expectation} for all $(x, y) \in \mathcal{X} \times \mathcal{Y}$.
  \STATE Obtain $\pi_{t+1}$ by Eq.~\eqref{eq:dno-objective}, where $\mathcal{D}_t$ is drawn as $x \sim \rho$, $y_1 \sim \mu_{1,t}(\cdot \mid x)$, $y_2 \sim \mu_{2,t}(\cdot \mid x)$ with $\mu_{1,t}, \mu_{2,t}$ off-policy or on-policy ($\pi_t$).
\ENDFOR
\STATE \textbf{return} $\bar{\pi} = \mathrm{unif}(\pi_{1:N})$
\end{algorithmic}
\end{algorithm}

\textbf{Guarantees:} Each exact iteration is one step of soft policy iteration, a no-regret update, so the uniform mixture $\bar{\pi}$ of the iterates converges to the Nash equilibrium of the two-player preference game, the Minimax Winner \citep{kreweras1965,simpson1969,kramer1973,fishburn1984}; the mixture argument is due to \citet{swamy2024spo}, and \citet{munos2023nlhf} establish last-iterate convergence for the regularized game. \citet{rosset2024dno} also establish two further properties (their Section~3.1 and Theorem~1). The exact soft-policy-iteration update improves monotonically, $\mathcal{P}(\pi^{\star}_{t+1} \succ \pi_t) \ge \tfrac12 + \eta\, \mathbb{E}_{x \sim \rho}[D_{\mathrm{KL}}(\pi^{\star}_{t+1} \,\|\, \pi_t)]$, and their Theorem~1 bounds the expected squared total-variation distance between the learned $\pi_{t+1}$ and $\pi^{\star}_{t+1}$ in terms of the sample size, the size of the policy class, a boundedness constant $R_{\max}$ and a data-dependent concentrability coefficient, so the improvement carries over up to that error. These statements hold under the assumptions of the cited results (realizability of the soft-policy-iteration target in the policy class, and boundedness), and the bound is informative only when the concentrability coefficient is small; neither condition is guaranteed for a neural policy class. The improvement is against the previous iterate under $\mathcal{P}$ and $\rho$; it is not a statement that benchmark scores rise at every iteration or that the student surpasses a fixed teacher. The practical procedure below, with finite samples, LLM ratings, margin filtering, look-back and a single selected checkpoint, approximates the exact update and does not inherit these guarantees.

\textbf{DNO-Prct:} The main obstacle in practice is the expectation in Eq.~\eqref{eq:reward-expectation}. \citet{rosset2024dno} therefore propose DNO-Prct, each iteration of which has three steps:
\begin{enumerate}[noitemsep,topsep=2pt,leftmargin=*]
    \item \textbf{Batched on-policy sampling and rating:} $K$ outputs per prompt are sampled from the current policy $\pi_t$ and, together with a gold response $\ygold$ from a stronger teacher model, are scored by an annotator; the scores serve as a proxy for the pairwise win-rate ranking under $\mathcal{P}$.
    \item \textbf{Preference pair construction:} Only pairs whose scores differ by a sufficiently large margin are kept, with the intent that the target $\sigma(\Delta^{\star}_t)$, after a rescaling of $r_t$, be close to one so that Eq.~\eqref{eq:dno-objective} reduces approximately to the pure contrastive (DPO) loss (an approximation discussed in Section~\ref{sec:ablation:errors}); a decaying fraction of pairs from earlier iterations is mixed in (data look-back, described in Section~5.1 of \citet{rosset2024dno} rather than in their Algorithm~2).
    \item \textbf{Contrastive update:} The policy is trained with the DPO objective on the constructed pairs, with $\pi_t$ as the reference policy.
\end{enumerate}
DNO-Prct is thus an iterative, on-policy form of DPO with the same contrastive objective; the difference lies in how the preference pairs are constructed (Figure~\ref{fig:pipeline} in Section~\ref{setup:dno-data}). We refer readers to \citet{rosset2024dno} for the derivation.

\section{Related Work}
\label{sec:related}

\textbf{RLHF and offline contrastive alignment:}
The reward-model-plus-RL recipe \citep{christiano2017deep,stiennon2020summarize,ouyang2022instructgpt,bai2022hh} is a standard route to aligned chat models. Offline contrastive methods such as DPO \citep{rafailov2023dpo}, IPO \citep{azar2023ipo} and SimPO \citep{meng2024simpo} fit the policy directly to a preference dataset; Zephyr \citep{tunstall2023zephyr} popularized distilled supervised fine-tuning (SFT) on UltraChat followed by DPO on UltraFeedback \citep{cui2023ultrafeedback}. DPO and SimPO inherit the transitivity of a pointwise (Bradley--Terry) reward, which IPO avoids, and DPO in particular is known to exploit response length \citep{park2024length}. \citet{meng2024simpo} report that Llama-3-8B-Instruct trained with SimPO reaches 44.7\% LC on the public AlpacaEval~2.0 leaderboard, above Llama-3-70B-Instruct (34.4\%), so a small model scoring above a larger sibling on this metric is not by itself evidence of a better algorithm.

\textbf{Iterative, online and self-play alignment:}
Iterative DPO \citep{xu2023cringe,xiong2024iterative,dong2024rlhfworkflow}, online AI feedback \citep{guo2024oaif} and Self-Rewarding LMs \citep{yuan2024selfrewarding} regenerate preference data from the current policy each round. \citet{tang2024online} find that the advantage of online over offline alignment is not explained by data coverage or quality alone and point to on-policy sampling as the deciding factor; \citet{lanchantin2025bridging} find that semi-online training, in which the response generator is synchronized with the policy only every few updates, as in DNO-Prct's per-iteration sampling, performs on par with fully online training and well above offline methods. SPIN \citep{chen2024spin} treats the SFT targets as always preferred over self-generated responses. Game-theoretic methods seek the Nash equilibrium of the preference game: Nash-MD \citep{munos2023nlhf}, online IPO \citep{calandriello2024online}, SPO \citep{swamy2024spo}, SPPO \citep{wu2024sppo}, INPO \citep{zhang2025inpo}, DNO \citep{rosset2024dno} and, concurrently with or after our experiments, MPO \citep{wang2025mpo}, optimistic online mirror descent \citep{zhang2025onpo} and RSPO \citep{tang2025rspo}. REBEL \citep{gao2024rebel} regresses relative rewards and extends to intransitive preferences, and \citet{zhang2025gpm} learn general preference models beyond Bradley--Terry for such methods. INPO reaches 42.6\% LC on AlpacaEval~2.0 from a Llama-3-8B SFT model with an open-weight pairwise preference model and no proprietary model in the loop. We have not found a published independent re-implementation of DNO-Prct on other base models, nor a public implementation of it.

\textbf{Where do the gains of AI feedback come from?}
\citet{sharma2024critical} show that improvements attributed to reinforcement learning from AI feedback (RLAIF) are largely explained by using a weaker teacher for SFT than the critic used for feedback, and that SFT directly on GPT-4 completions outperforms the pipelines they study. Our gold-response ablation (Section~\ref{sec:ablation:gold}) is related but not a replication of their comparison: we did not run a matched-teacher SFT control. \citet{ivison2024unpacking} disentangle preference data, learning algorithm, reward model and policy prompts in a comparable empirical study of DPO and PPO. Open-weight models trained specifically as evaluators can match or exceed GPT-4 on reward-model benchmarks \citep{wang2024selftaught}; our open-weight annotators were not evaluator-trained and retained 62--82\% of the LC gain of the GPT-4o-annotated runs against GPT-4o (Section~\ref{result:larger-annotators}).

\textbf{LLM-as-a-judge evaluation and its failure modes:}
We evaluate with AlpacaEval~2.0 \citep{li2023alpacaeval} and its length-controlled win rate \citep{dubois2024lcalpacaeval}, which with the default judge and reference correlates strongly with the rankings of Chatbot Arena \citep{chiang2024chatbotarena}; our judge and references are not the default (Section~\ref{setup:general-setup}). LLM judges exhibit position, verbosity and self-preference biases \citep{zheng2023judge,panickssery2024selfpref,wataoka2024selfpreference}. \citet{li2026preferenceleakage} find that a judge favors students trained on data from a related model: their average preference-leakage score is about 24\% when judge and data generator are the same model, as GPT-4o is in our main runs, and about 3--9\% when they share a model family. \citet{zheng2025nullmodels} show that a null model that always returns the same, adversarially crafted response reaches 86.5\% LC on AlpacaEval~2.0. Both appeared during or after our experiments, and Section~\ref{sec:limitations} uses them to interpret our results.

\subsection{Relation to Online DPO}
\label{sec:relation-with-online-dpo}
DNO is conceptually close to online DPO \citep{guo2024oaif}. In each round of online DPO the model generates two candidate responses to a prompt, an annotator decides which is better, and the resulting pair is used for a DPO-style update. DNO likewise samples candidates, rates them and learns from the resulting preferences, but differs in three ways.

\textbf{(1) Rating guideline and margin:} DNO rates responses against an explicit rubric and forms a pair only when the positive--negative rating margin is sufficiently large. The margin filter is what allows the regression objective of Eq.~\eqref{eq:dno-objective} to be approximated by the pure contrastive loss without an excessively large KL coefficient \citep[Section~4]{rosset2024dno}; it is an approximation device rather than a condition for Nash convergence. Its empirical effect is studied in Section~\ref{sec:ablation:margin}.

\textbf{(2) Multiple on-policy candidates:} Instead of a single pair per prompt, DNO samples $K$ candidates per prompt, intended to cover more of the policy's output distribution than a single pair. Section~\ref{sec:ablation:samples} varies $K$.

\textbf{(3) Teacher responses as candidates:} DNO adds the teacher's response $\ygold$ to the candidate set. Rating teacher and student responses together captures cases where the student outperforms the teacher, which SPIN-style self-play ignores (Sections~\ref{sec:ablation:gold} and~\ref{sec:ablation:pairs}).

Online DPO as instantiated by \citet{guo2024oaif} additionally keeps a fixed reference policy and samples pairs per mini-batch, whereas DNO-Prct collects a large batch per iteration, resets the reference to $\pi_t$, and mixes in look-back data.

\section{Experimental Setup}
\label{sec:setup}

\subsection{General Setup}
\label{setup:general-setup}

\textbf{Timeline:}
All experiments were run between the release of Llama~3.1 and Mistral Large~2 in late July 2024 and the end of December 2024, with the model and API versions available then. All reported results come from these 2024 runs; a few descriptive statistics, such as the prompt overlap and the truncation rate, were computed afterwards from their stored outputs.

\textbf{Dataset:}
We use the UltraFeedback dataset \citep{cui2023ultrafeedback,openbmb2023ultrafeedbackdata} as the fine-tuning and preference-alignment dataset. UltraFeedback comprises 63{,}967 prompts, four responses per prompt from a pool of seventeen models (including GPT-4, GPT-3.5 Turbo, Bard, Llama-2-70B-chat, Vicuna-33B and Falcon-40B-Instruct), and GPT-4 annotations along four aspects (instruction following, truthfulness, honesty and helpfulness), each on a 1--5 scale, together with an overall score on a 1--10 scale. The prompts are sourced from UltraChat, ShareGPT, Evol-Instruct, TruthfulQA, FalseQA and FLAN. The SFT and offline-DPO thresholds below are applied to the average of the four per-aspect ratings. In addition to the original content, we generate one response from GPT-4o \citep{openai2024gpt4o} for every prompt to serve as the gold response $\ygold$ in the GPT-4o-annotated runs.

\textbf{Base models and annotators:}
We experiment with two base models, Llama~3.1 8B Instruct and Mistral~7B Instruct~v0.1. The base model provides the initial weights for every run except the two-stage runs of Section~\ref{sec:results:two-stage} and the look-back ablation of Section~\ref{sec:ablation:pairs}, which start from an SFT checkpoint. GPT-4o is the annotator in the DNO experiments except those of Section~\ref{result:larger-annotators}, where the annotator is Llama~3.1 70B Instruct or Mistral Large~2 \citep[Mistral-Large-Instruct-2407, 123B parameters; hereafter Mistral Large;][]{mistral2024large}.

\textbf{Baselines:}
We compare DNO with our own implementations of supervised fine-tuning (SFT), offline DPO \citep{rafailov2023dpo}, a SPIN-style baseline \citep{chen2024spin}, DNO-Restrictive, SPPO \citep{wu2024sppo} and online DPO \citep{guo2024oaif,dong2024rlhfworkflow}. Our SPIN-style baseline marks the teacher's gold response as always preferred over the on-policy samples, which is the ``mark the teacher ($T$) always positive'' rule of Section~\ref{sec:ablation:pairs}; it is not the original SPIN recipe, which uses the SFT targets as winners. DNO-Restrictive accepts only pairs whose positive is the teacher's response. SPPO and online DPO were run on UltraFeedback with only the learning-rate search described below; both ended below the base model, and we therefore draw no conclusions about those methods from our runs.

\textbf{Pre-processing per algorithm:}
For SFT, we keep as targets the UltraFeedback responses whose average rating is five, the maximum, forming a question-answering dataset in OpenAI's conversation format with the ``user'' role for the prompt and the ``assistant'' role for the response. For offline DPO, we mark responses with an average rating of five as positive and responses rated at least two points lower as negative, and collect all positive--negative pairs per prompt; the prompt takes the ``user'' role followed by a generation token, and each chosen or rejected response is followed by an end-of-sequence or end-of-turn token. DNO uses only the prompts and the gold responses; on-policy responses are generated on the fly. We split the prompts into three disjoint partitions, one per DNO iteration, so that no prompt is used for new candidate generation in more than one iteration; pairs built in earlier iterations may be replayed through look-back (Section~\ref{setup:dno-data}).

\textbf{Training setup:}
We follow most hyperparameters of the Llama recipes \citep{meta2024llamarecipes}. The batch size is 2 per device with 4 gradient-accumulation steps on four NVIDIA A100-SXM 80\,GB GPUs, an effective batch size of 32 for all experiments, and training uses low-rank adaptation \citep[LoRA;][]{hu2022lora} (rank 16, $\alpha = 32$, dropout 0.05; Appendix~\ref{app:hparams}). We train SFT for 20 epochs (reporting the epoch-3 and epoch-10 checkpoints), DPO for 3 epochs, and DNO for three iterations, each a single epoch over that iteration's training pairs, to limit overfitting to them. Each iteration consists of on-policy generation, rating, large-margin pair selection and the contrastive update with the DPO loss ($\beta = 0.1$). It starts from the merged checkpoint of the previous iteration, and the reference policy is that checkpoint with the adapters disabled, i.e., $\pi_t$, as DNO-Prct prescribes. All algorithms are implemented with the Hugging Face TRL library \citep{vonwerra2020trl}. All runs use seed 2024.

\textbf{Learning rate:}
We search over $\{5\times10^{-5}, 5\times10^{-6}\}$ during the first epoch (first iteration for DNO) of every experiment and keep the better rate for the remaining epochs or iterations; the rate is $5\times10^{-5}$ for SFT and $5\times10^{-6}$ for DPO and most DNO experiments. The two rates differed by 5 to 10 points in win rate against GPT-4-0613, so in our runs the choice of learning rate had a large effect.

\textbf{Evaluation:}
One objective of this study is to see how far a small model can be pushed relative to a larger model of its family, so we evaluate all models, including the larger ones, under one pipeline: AlpacaEval~2.0 \citep[v0.6.5;][]{li2023alpacaeval} on its 805 prompts, reporting the raw and the length-controlled (LC) win rate \citep{dubois2024lcalpacaeval}. The protocol departs from the public leaderboard in three ways. (i)~Opponents: instead of \texttt{gpt4\_1106\_preview}, every model is compared with two opponents, GPT-4o and GPT-4-0613, whose reference outputs we generated through Azure OpenAI (temperature 0, at most 4{,}096 tokens). (ii)~Judge: the judge is GPT-4o, prompted with AlpacaEval's \texttt{alpaca\_eval\_gpt4} pairwise-ranking template at temperature 0, rather than the leaderboard's default GPT-4 Turbo annotator; GPT-4o is therefore also the teacher and one of the opponents (Section~\ref{sec:limitations}). (iii)~Decoding: each evaluated model answers with vLLM \citep{kwon2023vllm}, its chat template, temperature 0.001 and at most 4{,}096 new tokens. AlpacaEval computes the length-controlled win rate relative to whichever reference outputs it is given, so LC values are defined for both opponents, but neither LC nor raw values are comparable to the public leaderboard or to numbers in other papers, and the reported correlation of LC with Chatbot Arena (Spearman 0.98; \citealp{dubois2024lcalpacaeval}) was established for the default judge and reference, not for this protocol.

\textbf{Uncertainty:}
The standard error of the raw win rate reported by AlpacaEval is at most 1.9 points per evaluation (1.76 for the Llama Iteration-3 model against GPT-4o). Its standard error for the LC win rate (at most 0.26) is an error of the regression fit, not a sampling error, so we do not use it to compare models. These are evaluation uncertainties for single models; a comparison between two models on the shared prompts would need a paired analysis, which we have not done, and where we relate a difference to standard errors we treat the two evaluations as independent. The variation between training runs is unknown, because each configuration was trained once. We therefore read differences of a few points between reported runs as suggestive rather than established, and we do not claim that two configurations are equivalent when their scores are close.

\subsection{On-Policy Data Construction in DNO Iterations}
\label{setup:dno-data}
\begin{figure*}[t]
\centering
\includegraphics[width=\textwidth]{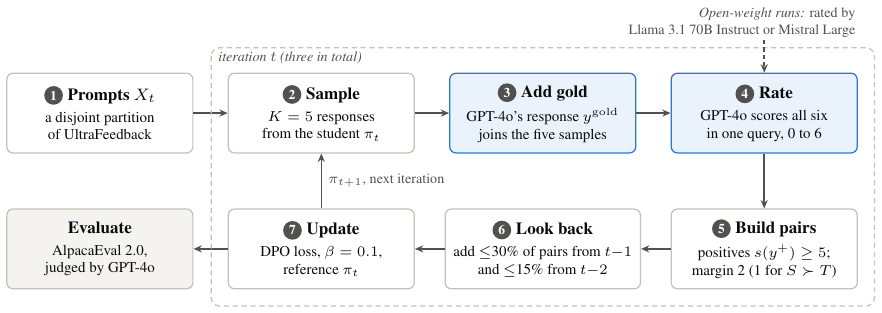}
\caption{One DNO-Prct iteration (Algorithm~\ref{alg:pairs}). Blue: training-time steps performed by GPT-4o in the main runs. Look-back pairs are reused without re-rating; LoRA adapters trained on $\pi_t$ are merged to give $\pi_{t+1}$.}
\label{fig:pipeline}
\end{figure*}

Throughout the paper we call the annotator the \emph{teacher} ($T$) and the model being fine-tuned the \emph{student} ($S$). In the GPT-4o runs the teacher also supplies the gold response $\ygold$. Figure~\ref{fig:pipeline} summarizes one iteration. In every block of experiments the candidates are sampled from the student $\pi_t$ and the evaluation judge is GPT-4o; in the open-weight-annotator runs (Section~\ref{result:larger-annotators}), Llama~3.1 70B Instruct or Mistral Large replaces GPT-4o as the annotator. A pair whose positive is the gold response $\ygold$ (the teacher side, $T$) and whose negative is a student response is written $\pTS$, and likewise $\pST$ and $\pSS$.

\textbf{On-policy response generation:}
At the start of each iteration we sample $K = 5$ responses per prompt (varied in Section~\ref{sec:ablation:samples}) from the current model, the initial checkpoint in the first iteration and the model from the previous iteration thereafter, with temperature 0.7 and nucleus sampling \citep[top-$p$, $p = 0.95$;][]{holtzman2020nucleus}.

\textbf{Rating:}
For the $K + 1$ responses per prompt ($K$ on-policy responses and $\ygold$; six in the main runs), the teacher rates quality following the preference-ranking approach of \citet{rosset2024dno}, with an additive rubric of six criteria whose integer scores range from 0 to 6 (Appendix~\ref{app:prompt}). Our prompt modifies the original, which was written for GPT-family annotators, so that it also works with Mistral Large and Llama~3.1 70B Instruct. Each response is evaluated along dimensions such as relevance, correctness and conciseness. For each prompt all candidates are sent to the annotator in one query, in a fixed order with $\ygold$ last, and scored side by side; the annotator returns one score per candidate. Candidates containing certain degenerate character sequences are dropped before rating, and a reply that cannot be parsed or does not contain one score per candidate is discarded.

\textbf{Preference pair construction:}
We select pairs $(x, y^{+}, y^{-})$ whose rating margin $\mathrm{rating}(y^{+}) - \mathrm{rating}(y^{-})$ is sufficiently large. We mark all responses with a score of at least five as positives. For each positive, responses rated at least two points lower are negatives, and we form all positive--negative pairs. If the positive is from the student and the negative is the teacher's response, we relax the margin to one point, because GPT-4o tends to rate its own responses highly \citep{panickssery2024selfpref,wataoka2024selfpreference}; whether such a gap reflects judge bias or a real quality difference cannot be determined from the scores alone \citep{chen2025evaluators}. This is a deviation from \citet{rosset2024dno}, who relax the margin whenever the positive is from the student, including $\pSS$ pairs, on the grounds that their teacher's responses rarely scored below five. Following the original paper, we include part of the data from past iterations to preserve earlier preference behavior (\emph{data look-back}): at iteration $t$ we sample at most 30\% of the training pairs of iteration $t-1$, 15\% of those of iteration $t-2$, and so on. These pairs are reused as they are, without re-rating. Appendix~\ref{app:pairs} gives the procedure as pseudo-code.

\section{Main Results}
\label{sec:results}
We report DNO and the baselines on two base models. Unless stated otherwise, GPT-4o is both the annotator and the source of $\ygold$. The DNO rows of Tables~\ref{result:main:llama} and~\ref{result:main:mistral} are the canonical runs that the direct-DNO numbers in this section refer to; the ablations of Section~\ref{sec:ablation} were run in a separate configuration and are compared with other runs only where stated; in particular, the SPIN-style and DNO-Restrictive rows of Table~\ref{result:main:llama} are those ablation runs, and the two-stage runs of Section~\ref{sec:results:two-stage} were trained in the ablation configuration.

\begin{table*}[t]
\centering
\small
\setlength{\tabcolsep}{6pt}
\caption{Llama~3.1 8B Instruct: DNO versus our reproduced baselines, with GPT-4o as annotator and gold-response source. AlpacaEval~2.0 LC / raw win rate (\%) against two opponents, judged by GPT-4o, and average length. The SPIN-style and DNO-Restrictive rows are ablation runs (Table~\ref{result:ablation:pair-selection}; full-DNO reference 43.90\% LC). Bold: best DNO value per column.}
\label{result:main:llama}
\begin{tabular}{llccc}
\toprule
Algorithm & \mc[l]{Epoch\\or iter.} & \mc{LC / raw win rate (\%)\\vs.\ GPT-4o} & \mc{LC / raw win rate (\%)\\vs.\ GPT-4-0613} & \mc{Avg.\ length\\(chars)} \\
\midrule
DNO & Iter.~1 & 39.01 / 43.04 & 61.44 / 68.45 & 3281 \\
DNO & Iter.~2 & 41.57 / 44.66 & 64.87 / 71.37 & 3626 \\
DNO & Iter.~3 & \textbf{45.32} / \textbf{48.26} & \textbf{70.18} / \textbf{74.60} & 3786 \\
\midrule
DNO-Restrictive & Iter.~3 & 37.98 / 38.62 & 60.16 / 60.51 & 4802 \\
SPIN-style & Epoch~3 & 35.17 / 41.99 & 50.30 / 66.81 & 3207 \\
SPPO & Epoch~3 & 31.28 / 34.22 & 45.85 / 59.44 & 2241 \\
Online DPO & Epoch~3 & 29.92 / 32.55 & 42.21 / 58.14 & 2076 \\
Offline DPO & Epoch~3 & 37.02 / 37.58 & 66.40 / 54.12 & 2173 \\
\midrule
SFT & Epoch~3 & 33.73 / 35.47 & 50.86 / 63.35 & 2417 \\
SFT & Epoch~10 & 24.01 / 23.98 & 39.64 / 49.88 & 2076 \\
\midrule
\mc[l]{Llama~3.1 8B Instruct\\(base model)} & -- & 31.71 / 35.51 & 49.64 / 62.19 & 2237 \\
\mc[l]{Llama~3.1 70B Instruct\\(larger model)} & -- & 41.40 / 39.94 & 55.23 / 67.64 & 1873 \\
\bottomrule
\end{tabular}
\end{table*}

\subsection{Llama~3.1 8B Instruct}
\label{sec:results:llama}
Table~\ref{result:main:llama} presents the main results with Llama~3.1 8B Instruct as the base model. DNO Iteration~3 performs best, with an LC win rate against GPT-4o that is 8.30 points above offline DPO (45.32\% vs.\ 37.02\%) and 11.59 points above SFT at epoch~3 (33.73\%). SFT degrades with more epochs (24.01\% at epoch~10, below the base model at 31.71\%); a plausible explanation, which we have not tested with a matched-data control, is that its targets are UltraFeedback responses from 2023-era models, many of them weaker than the instruct base model. A single DNO iteration (39.01\%) already scores 1.99 points above three epochs of offline DPO (37.02\%), and the second (41.57\%) is 4.55 points above it. Against GPT-4-0613 the three DNO iterations keep the same monotone ordering on both metrics, but offline DPO has a higher LC win rate (66.40\%) than DNO Iterations~1 and~2 (61.44\% and 64.87\%), and it is the only model in Table~\ref{result:main:llama} whose LC win rate against this opponent exceeds its raw win rate (54.12\%). The baselines are also not consistently ordered: offline DPO is above SFT (epoch~3) on LC win rate against GPT-4o (37.02\% vs.\ 33.73\%) but below it on raw win rate against GPT-4-0613 (54.12\% vs.\ 63.35\%).

Responses grow longer with every DNO iteration (3{,}281 to 3{,}786 characters on average). At Iteration~3 the median is 2{,}796 characters, and about 5\% of the responses reach or nearly reach the 4{,}096-token generation limit, so the average is raised by a small share of very long outputs, some of them truncated. The contrastive objective can widen the margin between positive and negative responses by lengthening outputs \citep{park2024length}, iterative DPO with preferences labeled by a trained reward model increases verbosity \citep{liu2024iterative}, and \citet[Section~5.2]{rosset2024dno} report a length increase of at least a factor of two in their own DNO runs. Because our rating step has no explicit length control, part of the raw-win-rate gain may be due to length. The LC win rate also rises with every iteration. This is consistent with more informative responses but does not establish it, because length control is a regression adjustment. For Mistral the length growth is far larger (Section~\ref{sec:results:mistral}); we return to length in Section~\ref{sec:limitations}.

\begin{table*}[t]
\centering
\small
\setlength{\tabcolsep}{5pt}
\caption{Mistral~7B Instruct~v0.1: DNO versus our reproduced baselines. Columns as in Table~\ref{result:main:llama}; the Rosset et al.\ column lists raw win rates from their own Mistral-based run, for reference only ($\sim$: approximate; see Section~\ref{sec:results:mistral}). Bold: best DNO value per column.}
\label{result:main:mistral}
\begin{tabular}{llcccc}
\toprule
Algorithm & \mc[l]{Epoch\\or iter.} & \mc{LC / raw win rate (\%)\\vs.\ GPT-4o} & \mc{LC / raw win rate (\%)\\vs.\ GPT-4-0613} & \mc{Reported by\\Rosset et al.\ (\%)} & \mc{Avg.\ length\\(chars)} \\
\midrule
DNO & Iter.~1 & 9.94 / 10.64 & 33.62 / 35.12 & $\sim$10 & 2424 \\
DNO & Iter.~2 & 13.62 / 13.45 & \textbf{38.67} / \textbf{40.17} & $\sim$21 & 5365 \\
DNO & Iter.~3 & \textbf{18.44} / \textbf{18.53} & 38.20 / 37.85 & 24.97 & 7506 \\
\midrule
Offline DPO & Epoch~3 & 7.05 / 5.85 & 23.27 / 24.63 & 18.22 & 1465 \\
SFT & Epoch~3 & 5.37 / 4.98 & 14.76 / 15.69 & 7.96 & 1804 \\
SFT & Epoch~10 & 4.28 / 4.11 & 13.77 / 14.82 & -- & 1808 \\
\midrule
\mc[l]{Mistral~7B Instruct~v0.1\\(base model)} & -- & 6.77 / 5.47 & 18.40 / 18.66 & -- & 1375 \\
\mc[l]{Mixtral~8x7B Instruct~v0.1\\(larger model)} & -- & 23.09 / 18.76 & 41.78 / 50.68 & -- & 1464 \\
\bottomrule
\end{tabular}
\end{table*}

\subsection{Mistral~7B Instruct~v0.1}
\label{sec:results:mistral}
Table~\ref{result:main:mistral} presents the comparison with Mistral~7B Instruct~v0.1. The raw win rates that \citet{rosset2024dno} report for their Mistral-based run are listed alongside for reference, but they are not comparable with ours: they were measured against GPT-4 Turbo with a GPT-4 Turbo annotator, starting from Orca-2.5, an SFT variant of Mistral~7B. Their Iteration~3, offline DPO and SFT-on-positives values are from their Table~1, where the baselines are reported at epoch~4; the values for Iterations~1 and~2, which their tables do not list, are approximate. As with Llama~3.1, the LC win rate against GPT-4o rises with every iteration, and Iteration~3 scores above offline DPO by 11.39 points and above SFT by 13.07 points in LC win rate against GPT-4o (18.44\% vs.\ 7.05\% and 5.37\%), an 11.67-point gain over the base model. Against GPT-4-0613, both metrics decline slightly after Iteration~2 (LC 38.67\% to 38.20\%; raw 40.17\% to 37.85\%), whereas against GPT-4o they keep rising; whether this is saturation or run noise cannot be determined from single runs. One untested hypothesis is that the GPT-4o gold responses pull the student toward GPT-4o-like outputs, which the judge may score differently when the opponent is GPT-4o itself than when it is GPT-4-0613. The average response length grows to 7{,}506 characters, more than five times the base model's.

\subsection{Comparison with Larger Models of the Same Family}
\label{sec:results:larger}
To probe the limits of the student, we compare each DNO-tuned model with a larger model from the same family and release period, evaluated under the identical pipeline: Llama~3.1 70B Instruct and Mixtral~8x7B Instruct~v0.1, respectively. In Table~\ref{result:main:llama}, Llama~3.1 8B Instruct after three DNO iterations scores above its 70B variant, which has nearly nine times as many parameters, on all four metrics; against GPT-4-0613 the LC difference is 14.95 points (70.18\% vs.\ 55.23\%), and against GPT-4o it is 3.92 points (45.32\% vs.\ 41.40\%). The raw win rates against GPT-4o differ by 8.32 points (48.26\% vs.\ 39.94\%, standard errors 1.76 and 1.72), about 3.4 standard errors of the difference between two independent evaluations. Two qualifications apply: the two models were not compared head-to-head but each against the same opponents; and GPT-4o, which produced the gold responses and rated every candidate, is also one of those opponents and the judge. The GPT-4-0613 column uses an opponent from an earlier model generation, but the judge is still GPT-4o, so it does not remove this exposure. Mistral~7B Instruct~v0.1 with DNO (Iteration~3) comes close to Mixtral~8x7B on raw win rate against GPT-4o (18.53\% vs.\ 18.76\%) but remains behind on LC win rate against GPT-4o (18.44\% vs.\ 23.09\%) and on both metrics against GPT-4-0613 (38.20\% vs.\ 41.78\% LC; 37.85\% vs.\ 50.68\% raw). Mixtral has roughly 6.5 times the total parameters of the 7B model (46.7B vs.\ 7.2B; 12.9B active per token) in a mixture-of-experts architecture; whether that, the base model's capacity, or the data accounts for the remaining gap we cannot tell from these runs.

\begin{table*}[tp]
\centering
\scriptsize
\setlength{\tabcolsep}{2.6pt}
\caption{Llama~3.1 8B Instruct under three DNO settings: direct DNO annotated by GPT-4o (the run of Table~\ref{result:main:llama}), direct DNO annotated by Llama~3.1 70B Instruct, and SFT followed by DNO annotated by GPT-4o. Cells: LC / raw win rate (\%); Len.: average length (chars). Bold: best DNO value per column within each block.}
\label{result:diff-annotators:llama}
\begin{tabular}{lccccccccc}
\toprule
\multirow{2}{*}{Algorithm} & \multicolumn{3}{c}{Direct DNO, annotator GPT-4o} & \multicolumn{3}{c}{Direct DNO, annotator Llama~3.1 70B} & \multicolumn{3}{c}{SFT + DNO, annotator GPT-4o} \\
\cmidrule(lr){2-4}\cmidrule(lr){5-7}\cmidrule(lr){8-10}
 & vs.\ GPT-4o & vs.\ GPT-4-0613 & Len. & vs.\ GPT-4o & vs.\ GPT-4-0613 & Len. & vs.\ GPT-4o & vs.\ GPT-4-0613 & Len. \\
\midrule
DNO Iter.~1 & 39.01 / 43.04 & 61.44 / 68.45 & 3281 & 37.35 / 44.34 & 56.95 / 71.08 & 2763 & 36.85 / 38.20 & 50.52 / 65.53 & 2092 \\
DNO Iter.~2 & 41.57 / 44.66 & 64.87 / 71.37 & 3626 & 40.01 / \textbf{47.45} & 58.55 / \textbf{72.55} & 3025 & 38.90 / 44.22 & 55.06 / \textbf{72.42} & 2285 \\
DNO Iter.~3 & \textbf{45.32} / \textbf{48.26} & \textbf{70.18} / \textbf{74.60} & 3786 & \textbf{40.13} / \textbf{47.45} & \textbf{58.91} / 71.80 & 3185 & \textbf{44.92} / \textbf{45.14} & \textbf{68.96} / 69.30 & 4342 \\
\midrule
Llama~3.1 8B Instruct (base) & 31.71 / 35.51 & 49.64 / 62.19 & 2237 & -- & -- & -- & -- & -- & -- \\
Llama~3.1 70B Instruct (larger) & 41.40 / 39.94 & 55.23 / 67.64 & 1873 & -- & -- & -- & -- & -- & -- \\
\bottomrule
\end{tabular}
\end{table*}

\begin{table*}[tp]
\centering
\scriptsize
\setlength{\tabcolsep}{2.6pt}
\caption{Mistral~7B Instruct~v0.1 under the three DNO settings of Table~\ref{result:diff-annotators:llama}, with Mistral Large as the open-weight annotator.}
\label{result:diff-annotators:mistral}
\begin{tabular}{lccccccccc}
\toprule
\multirow{2}{*}{Algorithm} & \multicolumn{3}{c}{Direct DNO, annotator GPT-4o} & \multicolumn{3}{c}{Direct DNO, annotator Mistral Large} & \multicolumn{3}{c}{SFT + DNO, annotator GPT-4o} \\
\cmidrule(lr){2-4}\cmidrule(lr){5-7}\cmidrule(lr){8-10}
 & vs.\ GPT-4o & vs.\ GPT-4-0613 & Len. & vs.\ GPT-4o & vs.\ GPT-4-0613 & Len. & vs.\ GPT-4o & vs.\ GPT-4-0613 & Len. \\
\midrule
DNO Iter.~1 & 9.94 / 10.64 & 33.62 / 35.12 & 2424 & \textbf{16.40} / 17.36 & 38.72 / \textbf{40.21} & 8264 & 10.18 / 10.43 & \textbf{30.83} / 34.53 & 2078 \\
DNO Iter.~2 & 13.62 / 13.45 & \textbf{38.67} / \textbf{40.17} & 5365 & 14.29 / 15.50 & \textbf{39.16} / 38.81 & 5365 & 11.43 / 10.68 & 29.29 / 32.92 & 1768 \\
DNO Iter.~3 & \textbf{18.44} / \textbf{18.53} & 38.20 / 37.85 & 7506 & 16.30 / \textbf{17.81} & 38.71 / 39.09 & 7699 & \textbf{12.58} / \textbf{13.29} & 30.56 / \textbf{37.27} & 2030 \\
\midrule
Mistral~7B Instruct~v0.1 (base) & 6.77 / 5.47 & 18.40 / 18.66 & 1375 & -- & -- & -- & -- & -- & -- \\
Mixtral~8x7B Instruct~v0.1 (larger) & 23.09 / 18.76 & 41.78 / 50.68 & 1464 & -- & -- & -- & -- & -- & -- \\
\bottomrule
\end{tabular}
\end{table*}

\subsection{Open-Weight Annotators from the Same Model Family}
\label{result:larger-annotators}
In this section we switch the annotator from the proprietary GPT-4o to a larger open-weight model from the student's own family: Llama~3.1 70B Instruct annotates for Llama~3.1 8B Instruct, and Mistral Large annotates for Mistral~7B Instruct~v0.1. These experiments show that the rating step of DNO can be run with an open-weight model; we make this claim for the rating step only, because the source of $\ygold$ in these runs was not recorded. In this setting DNO can also be read as preference distillation from the larger model: the student learns from the teacher's preference signal and, where $\ygold$ comes from the teacher, from one of its responses.

The left and middle blocks of Table~\ref{result:diff-annotators:llama} compare the two annotators for Llama. Raw win rates against GPT-4o are similar (47.45\% with the Llama annotator vs.\ 48.26\% with GPT-4o), while the LC win rate is 5.19 points lower (40.13\% vs.\ 45.32\%); against GPT-4-0613 the LC gap is 11.27 points (58.91\% vs.\ 70.18\%), which may reflect more accurate ratings from GPT-4o or a judge preference for GPT-4o-like outputs (Section~\ref{sec:limitations}). Relative to the base model, the run with the open-weight annotator retains 8.4 of the 13.6 LC points gained against GPT-4o (62\%) and 9.3 of the 20.5 points gained against GPT-4-0613 (45\%). Even with this annotator, the 8B student scores above the 70B model on three of four metrics and 1.27 LC points below it against GPT-4o.

With Mistral Large as teacher (middle block of Table~\ref{result:diff-annotators:mistral}), the student improves over the base model: the LC win rate rises from 6.77\% to 16.30\% against GPT-4o (9.5 of the 11.7 points gained with GPT-4o as annotator, 82\%) and from 18.40\% to 38.71\% against GPT-4-0613, matching the GPT-4o-annotated run on that metric (38.20\%; 0.5 points higher). Unlike the GPT-4o-annotated runs, the LC win rate against GPT-4o reaches its highest value in the first iteration (16.40\%) and does not exceed it afterwards (14.29\% and 16.30\%), and against GPT-4-0613 it stays between 38.71\% and 39.16\%; response length varies widely across iterations (8{,}264, 5{,}365 and 7{,}699 characters). On inspection, the annotations from Mistral Large were less consistent in formatting than those of GPT-4o or Llama~3.1 70B Instruct; we have not quantified the rating-failure rate or the number of usable pairs per iteration, so we do not attribute the flat trajectory to a specific cause.

\subsection{Two-Stage Training: SFT Followed by DNO}
\label{sec:results:two-stage}
To compare with the conventional two-stage recipe \citep{ouyang2022instructgpt,tunstall2023zephyr}, we first fine-tune each base model with SFT and then apply DNO on UltraFeedback, with the pre-processing of Section~\ref{setup:general-setup}; DNO then starts from the SFT checkpoint.

The right-hand blocks of Tables~\ref{result:diff-annotators:llama} and~\ref{result:diff-annotators:mistral} compare two-stage training with direct DNO on the instruct model. For Llama the two recipes end close to each other in LC win rate (44.92\% vs.\ 45.32\% against GPT-4o; 68.96\% vs.\ 70.18\% against GPT-4-0613), while direct DNO keeps a lead in raw win rate (3.1 and 5.3 points). These comparisons use the canonical run of Table~\ref{result:main:llama}; the two-stage runs were trained in the configuration of the ablation reference of Table~\ref{result:ablation:pair-selection} (43.90\% and 69.46\%), against which two-stage training is 1.0 point higher on LC win rate against GPT-4o and 0.5 points lower against GPT-4-0613. Two-stage responses are shorter than direct DNO's in the first two iterations (2{,}092 and 2{,}285 vs.\ 3{,}281 and 3{,}626 characters) but longer at Iteration~3 (4{,}342 vs.\ 3{,}786); in both recipes the LC win rate peaks at the third iteration. For Mistral the conclusion does not transfer: two-stage training has lower point estimates on all four metrics (LC 12.58\% vs.\ 18.44\% against GPT-4o and 30.56\% vs.\ 38.20\% against GPT-4-0613; raw 13.29\% vs.\ 18.53\% and 37.27\% vs.\ 37.85\%), and its responses are about 73\% shorter at Iteration~3 (2{,}030 vs.\ 7{,}506 characters). The SFT stage is a plausible cause: the epoch-3 SFT checkpoint lowers Mistral's LC win rate against GPT-4-0613 by 3.6 points (14.76\% vs.\ 18.40\% in Table~\ref{result:main:mistral}; by 1.4 points against GPT-4o). We have not tested this explanation with a different SFT initialization.

\section{Ablation Studies and Analysis}
\label{sec:ablation}
All ablations use Llama~3.1 8B Instruct with GPT-4o as annotator and three iterations, in configurations separate from the canonical runs, and ablation runs are compared with runs outside their own table only where stated. The full-DNO reference row of Tables~\ref{result:ablation:num_samples} and~\ref{result:ablation:pair-selection} (43.90\% LC against GPT-4o, 5{,}324 characters) is a different run from the canonical run of Table~\ref{result:main:llama} (45.32\%, 3{,}786 characters). The margin sweep of Table~\ref{result:ablation:margin} was run with its own pair-construction and training scripts and at most 2{,}000 training steps per iteration. Within each sweep, the hyperparameters other than the swept one are fixed, but the number and mix of selected pairs, and hence of updates, change with it, so a sweep does not isolate one factor.

\begin{table*}[tp]
\centering
\small
\setlength{\tabcolsep}{6pt}
\caption{Number of on-policy samples per prompt (ablation configuration). Selected pairs by type ($T$: teacher's gold response; $S$: student response); ratings on the 0--6 scale.}
\label{result:ablation:num_samples}
\begin{tabular}{cccrrrrc}
\toprule
\mc{Samples\\per prompt} & \mc{LC / raw win rate (\%)\\vs.\ GPT-4o} & \mc{Avg.\ length\\(chars)} & $\pTS$ & $\pST$ & $\pSS$ & Total & \mc{Avg.\ rating\\pos.\ / neg.} \\
\midrule
1 & 44.90 / 39.68 & 3251 & 4{,}025  & 2{,}708  & 0      & 6{,}733  & 5.87 / 3.39 \\
3 & 45.96 / 41.37 & 2700 & 14{,}984 & 5{,}569  & 7{,}793  & 28{,}346 & 5.84 / 3.41 \\
5 & 43.90 / 41.77 & 5324 & 20{,}978 & 14{,}167 & 33{,}379 & 68{,}524 & 5.92 / 3.44 \\
7 & 44.15 / 38.75 & 3116 & 14{,}145 & 9{,}060  & 45{,}614 & 68{,}819 & 5.79 / 3.35 \\
\bottomrule
\end{tabular}
\end{table*}

\subsection{Number of On-Policy Samples}
\label{sec:ablation:samples}
We vary the number of on-policy samples per prompt over 1, 3, 5 and 7 (Table~\ref{result:ablation:num_samples}). The LC values lie within a 2.1-point range (43.90\% to 45.96\%) with no trend in the number of samples, so within this range the sample count changes the LC win rate little; the main experiments use five samples, the default of \citet{rosset2024dno}. Two observations hold regardless. First, the settings differ in more than the sample count: the reported number of selected pairs spans an order of magnitude (6{,}733 for one sample; 68{,}524 for five; 68{,}819 for seven), and with one epoch per iteration more pairs mean more updates. Second, the average ratings of positives and negatives are similar across settings (5.8--5.9 and 3.4); averages do not capture the diversity or coverage of the selected pairs, which we did not measure.

\begin{table*}[tp]
\centering
\small
\setlength{\tabcolsep}{5pt}
\caption{Minimum rating margin (separate pipeline, at most 2{,}000 training steps per iteration). Columns as in Table~\ref{result:ablation:num_samples}, plus the share of pairs whose positive is the teacher's response.}
\label{result:ablation:margin}
\begin{tabular}{cccrrrrcc}
\toprule
Margin & \mc{LC / raw win rate (\%)\\vs.\ GPT-4o} & \mc{Avg.\ length\\(chars)} & $\pTS$ & $\pST$ & $\pSS$ & Total & \mc{Share of pairs with\\teacher positive (\%)} & \mc{Avg.\ rating\\pos.\ / neg.} \\
\midrule
1 & 36.60 / 41.23 & 2978 & 45{,}344 & 24{,}643 & 78{,}777 & 148{,}764 & 30.48 & 5.78 / 3.70 \\
2 & 37.72 / 42.92 & 2931 & 16{,}601 & 13{,}089 & 25{,}712 & 55{,}402  & 29.96 & 5.81 / 3.39 \\
3 & 38.99 / 43.54 & 2619 & 9{,}377  & 6{,}784  & 8{,}765  & 24{,}926  & 37.62 & 5.79 / 2.01 \\
4 & 37.38 / 40.98 & 2551 & 5{,}169  & 4{,}580  & 3{,}826  & 13{,}575  & 38.08 & 5.82 / 1.42 \\
5 & 38.13 / 41.93 & 2542 & 638     & 923     & 254     & 1{,}815   & 35.15 & 5.97 / 0.92 \\
\bottomrule
\end{tabular}
\end{table*}

\subsection{Rating Margin}
\label{sec:ablation:margin}
A large calibrated reward gap motivates the contrastive approximation to Eq.~\eqref{eq:dno-objective}; a large rubric-score gap is only a proxy for that condition (Section~\ref{sec:ablation:errors}), but a margin that is too large discards most pairs and shifts the data mix toward teacher positives. Table~\ref{result:ablation:margin} evaluates margins of one to five with all other hyperparameters fixed; the number and mix of selected pairs, and hence of updates, change with the margin. The LC win rate against GPT-4o lies between 36.60\% and 38.99\% for all five margins, with the highest value at margin three; over this range the LC win rate changes by only 2.4 points, and the main experiments keep the default margin of two from \citet{rosset2024dno}. \citet[Section~5.2]{rosset2024dno} compared margin filtering with training on all pairs and found filtering beneficial; our sweep varies the margin but has no unfiltered run. Three observations follow from the pair statistics. First, the share of pairs whose positive is the teacher's response is about 30\% at margins one and two and 35--38\% at margins three to five; it is not monotonic in the margin. Second, the average rating of positives stays at 5.8--6.0 while that of negatives falls from 3.70 to 0.92, so large margins select very poor negatives. We did not measure whether such pairs are less informative per sample. Under the DPO loss the per-pair gradient weight is $\sigma(-z)$, where $z$ is the model's implicit log-ratio margin scaled by $\beta$; because the reference policy is the starting checkpoint of each iteration (Section~\ref{setup:general-setup}), $z = 0$ for every pair at the first update with the default adapter initialization, regardless of the rubric gap, so the rubric gap does not by itself change the initial gradient weight. Third, the average response length decreases with the margin, which may partly reflect the larger share of the teacher's typically shorter responses at margins three to five, although the share does not track length at every step. The margin also sets the data volume, from 148{,}764 reported selected pairs at margin one to 1{,}815 at margin five.

\textbf{Comparability with the other ablations:} All five margin runs score below the full-DNO reference of Table~\ref{result:ablation:pair-selection} (43.90\% LC against GPT-4o), including the margin-2 run (37.72\%), which uses the same margin. Because the sweep used a separate pipeline (Section~\ref{sec:ablation}), this gap is confounded with implementation differences that single runs cannot separate from run-to-run variation, so we compare the margin runs only with each other.

\begin{table*}[tp]
\centering
\small
\setlength{\tabcolsep}{6pt}
\caption{Gold responses, pair-selection rule and data look-back (ablation configuration). $\Delta$LC: change relative to full DNO. ``Mark $T$ always positive'' is the SPIN-style rule and ``Accept only $\pTS$'' the DNO-Restrictive rule. The look-back variant starts from the SFT checkpoint; its matched comparison is given in Section~\ref{sec:ablation:pairs}. Bold: best value per win-rate column.}
\label{result:ablation:pair-selection}
\begin{tabular}{lrrrrrrc}
\toprule
\multirow{2}{*}{Variant} & \multicolumn{3}{c}{vs.\ GPT-4o (\%)} & \multicolumn{3}{c}{vs.\ GPT-4-0613 (\%)} & \mc{Avg.\ length\\(chars)} \\
\cmidrule(lr){2-4}\cmidrule(lr){5-7}
 & LC & $\Delta$LC & raw & LC & $\Delta$LC & raw & \\
\midrule
DNO (full) & \textbf{43.90} & -- & 41.77 & \textbf{69.46} & -- & 64.21 & 5324 \\
\quad Remove $\ygold$ & 36.97 & $-$6.93 & 40.91 & 55.36 & $-$14.10 & 66.15 & 2648 \\
\quad Mark $T$ always positive & 35.17 & $-$8.73 & \textbf{41.99} & 50.30 & $-$19.16 & 66.81 & 3207 \\
\quad Accept only $\pTS$ & 37.98 & $-$5.92 & 38.62 & 60.16 & $-$9.30 & 60.51 & 4802 \\
\quad Remove data look-back & 35.00 & $-$8.90 & 41.61 & 50.10 & $-$19.36 & \textbf{67.45} & 2489 \\
\bottomrule
\end{tabular}
\end{table*}

\subsection{Gold Responses}
\label{sec:ablation:gold}
We examine the role of $\ygold$. In addition to preference signals, DNO includes the teacher's response as one of the candidates; in practice, gold responses may be unavailable because they cost more than preference labels alone. Removing them (row ``Remove $\ygold$'' of Table~\ref{result:ablation:pair-selection}) lowers the LC win rate against GPT-4o from 43.90\% to 36.97\% and against GPT-4-0613 from 69.46\% to 55.36\%, while the raw win rates change by $-$0.9 and $+$1.9 points. The drop is consistent with the teacher's responses being rated well above the student's in the first iteration (Table~\ref{result:ablation:teacher-student-evolution}). Without $\ygold$, DNO becomes a bootstrapped online DPO with more than two candidates per prompt. Its LC win rate against GPT-4o (36.97\%) is close to that of our offline DPO baseline (37.02\% in Table~\ref{result:main:llama}); the two runs use different data and training budgets, so the proximity is not a controlled comparison. Two controls that we did not run would separate the questions this raises: (i) offline DPO under the ablation training budget, to test whether the budget explains the proximity; and (ii) offline DPO trained from the same initialization on the Iteration-1 DNO pairs, with matched update budget, pair rule, gold inclusion and prompt coverage, to test whether iterative on-policy resampling adds anything to a single round of teacher-rated data. With preference signals alone, the variant without $\ygold$ still scores 5.3 LC points above the base model (31.71\%).

\subsection{Pair-Selection Rule and Data Look-Back}
\label{sec:ablation:pairs}
Table~\ref{result:ablation:pair-selection} compares the full pair-selection rule with two restricted rules. Always marking the teacher's response as positive and all student responses as negative (the SPIN-style rule) lowers the LC win rate by 8.73 points against GPT-4o and by 19.16 points against GPT-4-0613. A plausible reason is that this rule labels a student response as negative even when the annotator rates it above the teacher's, which becomes more frequent in later iterations (Table~\ref{result:ablation:teacher-student-evolution}). Accepting only $\pTS$ pairs, i.e., trusting the teacher only when the rating confirms it (DNO-Restrictive), lands between the two against GPT-4o ($-$5.92 vs.\ $-$8.73 points) and about half as far below the full rule against GPT-4-0613 ($-$9.30 vs.\ $-$19.16); the full rule scores highest on both LC metrics. On raw win rate the SPIN-style rule is level with or above the full rule, while DNO-Restrictive is 3 to 4 points below; the benefit of the full rule therefore shows mainly in the length-controlled metric, and the full-DNO reference also has the longest responses in the table (5{,}324 characters vs.\ 2{,}489 to 4{,}802), so this benefit depends on the length adjustment, which we have not validated for responses this long (Section~\ref{sec:limitations}). On LC win rate our ordering, SPIN-style $<$ DNO-Restrictive $<$ full rule, matches the one \citet[Table~1]{rosset2024dno} report in their setting.

Removing the training pairs carried over from previous iterations (row ``Remove data look-back'') lowers the LC win rate by 8.90 and 19.36 points relative to the full-DNO reference, while the raw win rates are almost unchanged against GPT-4o (41.61\% vs.\ 41.77\%) and higher against GPT-4-0613 (67.45\% vs.\ 64.21\%). This variant was trained from the SFT checkpoint, so its like-for-like counterpart is the two-stage Iteration-3 run of the same run family, trained with look-back (Table~\ref{result:diff-annotators:llama}: 44.92\% / 45.14\% against GPT-4o and 68.96\% / 69.30\% against GPT-4-0613). Against that run, removing look-back lowers the LC win rate by 9.92 and 18.86 points and the raw win rate by 3.53 and 1.85 points, so the decline holds on all four metrics. One hypothesis is that, because candidate responses are generated on-policy, a model that no longer sees earlier pairs regresses to behaviors that were already penalized; we observed only the aggregate scores, not those behaviors.

Against GPT-4o, all four variants are 5.9 to 8.9 LC points below the full-DNO reference, and the look-back variant is 9.9 points below its matched two-stage counterpart. These differences do not apportion the gain among the components.

\begin{table*}[tp]
\centering
\small
\setlength{\tabcolsep}{4pt}
\caption{Teacher ($T$, GPT-4o gold response) versus student ($S$) across DNO iterations, with GPT-4o as annotator. Row ``Iter.~$t$'' describes the responses sampled from $\pi_t$ at the start of iteration $t$, on that iteration's prompt partition (Section~\ref{setup:general-setup}); pair counts are new pairs only, excluding look-back.}
\label{result:ablation:teacher-student-evolution}
\begin{tabular}{lccrrrrcc}
\toprule
 & \mc{Avg.\ length $T$ / $S$\\(chars)} & \mc{Avg.\ rating\\$T$ / $S$} & $\pTS$ & $\pST$ & $\pSS$ & Total & \mc{Share of pairs with\\student positive (\%)} & \mc{Avg.\ rating\\pos.\ / neg.} \\
\midrule
Llama~3.1 8B, Iter.~1   & 1116.45 / 1521.30 & 4.86 / 3.28 & 25{,}093 & 2{,}373  & 21{,}543 & 49{,}009 & 48.80 & 5.74 / 1.68 \\
Llama~3.1 8B, Iter.~2   & 1121.66 / 1744.86 & 4.42 / 4.43 & 12{,}852 & 11{,}342 & 14{,}058 & 38{,}252 & 66.40 & 5.62 / 2.70 \\
Llama~3.1 8B, Iter.~3   & 1103.87 / 2003.26 & 4.26 / 4.52 & 6{,}057  & 7{,}602  & 7{,}747  & 21{,}406 & 71.70 & 5.62 / 2.67 \\
\midrule
Mistral~7B, Iter.~1     & 1147.87 / \phantom{0}806.08 & 5.03 / 3.48 & 42{,}576 & 2{,}273  & 19{,}129 & 63{,}978 & 33.45 & 5.69 / 2.60 \\
Mistral~7B, Iter.~2     & 1252.91 / 2048.87 & 4.63 / 3.93 & 20{,}450 & 5{,}518  & 24{,}943 & 50{,}911 & 59.83 & 5.61 / 2.51 \\
Mistral~7B, Iter.~3     & 1446.46 / 2951.22 & 4.66 / 4.13 & 6{,}268  & 2{,}581  & 18{,}412 & 27{,}261 & 77.01 & 5.58 / 3.07 \\
\bottomrule
\end{tabular}
\end{table*}

\subsection{Student Evolution under the Teacher--Student Framework}
\label{sec:teacher-student-framework}
We examine how the rated quality of the student's responses evolves relative to the teacher's, repeating on two other base models the per-iteration analysis of \citet[Table~2]{rosset2024dno}, who used a GPT-4 Turbo teacher. Table~\ref{result:ablation:teacher-student-evolution} compares teacher and student responses on length, rating and pair composition; each row describes the responses sampled at the start of that iteration, not the checkpoint at its end. The student's average rating rises with every iteration, and so does the share of selected pairs whose positive is a student response. For Llama~3.1 8B Instruct, the student's average rating in Iteration~1 is 3.28, 1.58 points below the teacher's 4.86; the two averages are nearly equal in Iteration~2 (4.43 vs.\ 4.42, on a different prompt partition) and the student's is 0.26 points higher in Iteration~3. By Iteration~3 there are also more $\pST$ than $\pTS$ pairs (7{,}602 vs.\ 6{,}057), showing that the selection rule admits student-positive, teacher-negative pairs, as \citet{rosset2024dno} also observed; this count is not a test of equilibrium attainment or teacher-level performance. Mistral~7B Instruct~v0.1 shows the same trend: the student's rating improves across iterations and its share of pairs with a student positive more than doubles, from 33.45\% to 77.01\%, although its rating (4.13) stays below the teacher's (4.66) in these runs.

Three caveats qualify this picture. First, the comparison is made on the annotator's own rubric: the same GPT-4o that produced $\ygold$ also scores it, and training-time rubric scores and external benchmark win rates measure different things. Under AlpacaEval the Llama Iteration-3 model wins 388, loses 416 and ties one of the 805 comparisons with GPT-4o (raw win rate 48.26\%, standard error 1.76), which is not distinguishable from parity. Second, the teacher's average rating falls across iterations (4.86 to 4.26 for Llama) even though its responses are fixed, as \citet{rosset2024dno} also report for their teacher; because each iteration uses a different prompt partition and the six responses are scored side by side, we cannot separate harder prompts from relative scoring, and the change in the gap should not be read as a decomposition of student improvement versus rating drift. Third, the counts are shaped by the pair-selection rule: five student samples face one teacher response, and the one-point relaxation of the margin for $\pST$ pairs inflates their count relative to $\pTS$ pairs. Within these caveats, the results support letting a student response serve as the positive when the annotator rates it above the teacher's, instead of always preferring the teacher as the SPIN-style rule does (Section~\ref{sec:ablation:pairs}).

\subsection{Sources of Approximation Error}
\label{sec:ablation:errors}
The experiments above expose several places where DNO-Prct departs from the theory of Section~\ref{sec:dno}.

\textbf{The large-margin approximation:}
Since $r_t(x, y) \in [0, 1]$, the target weight $\sigma(r_t(x, y^{+}) - r_t(x, y^{-}))$ is at most $\sigma(1) \approx 0.73$ and cannot approach one without rescaling. \citet{rosset2024dno} therefore rescale $r_t \to c\, r_t$ and $\eta \to c\, \eta =: \tilde{\eta}$, which leaves the soft-policy-iteration target unchanged, and note that the appropriate $\tilde{\eta}$, the coefficient of the log-ratio inside the contrastive loss (DPO's $\beta$), is thus expected to be larger than $\eta$; because a larger $\tilde{\eta}$ worsens the sample complexity of their Theorem~1, only large-margin pairs are retained so that $\tilde{\eta}$ need not be too large. Two points follow. First, although \citet{rosset2024dno} call $\tilde{\eta}$ a learning rate, it is the coefficient of the log-ratio, not the optimizer's learning rate; our learning-rate search does not test this scaling argument, and our runs use $\beta = 0.1$ (Appendix~\ref{app:hparams}). A sweep over $\beta$ at a fixed margin is the experiment that would test it. Second, a large gap on an ordinal 0--6 rating scale is not a calibrated large gap in the theoretical reward $r_t$, so the margin filter is a modeling approximation whose validity we have not measured; in the margin sweep above, margins of one to five changed the LC win rate by at most 2.4 points at this step budget.

\textbf{Rating noise:}
Since we use an LLM to grade responses in a batched setting, following \citet{rosset2024dno}, the ratings depend on the annotator's instruction following \citep{zheng2023judge}. Presenting all candidates in one query relies on the annotator to score each response separately and consistently: in GPT-4o's rating replies we observed incomplete ratings, where fewer scores are returned than candidates, and inconsistencies in the indexing of the scores; Mistral Large was less consistent still (Section~\ref{result:larger-annotators}). Such replies are discarded, so the affected prompts contribute no pairs. The candidates are presented in a fixed order with $\ygold$ last, so any position bias of the annotator \citep{zheng2023judge} falls systematically on the gold response. The ratings have no explicit length control, and the student's responses are longer than the teacher's from the first iteration for Llama and from the second for Mistral (Table~\ref{result:ablation:teacher-student-evolution}).

\textbf{Finite on-policy samples:}
The on-policy sample (Step~1 of DNO-Prct) is drawn once from $\pi_t$ at the start of the iteration, so it becomes increasingly off-policy as the update proceeds, and five samples per prompt cover only a small part of the policy's output distribution; we did not measure their diversity.

\section{Costs of DNO}
\label{sec:costs}
We report the cost of a three-iteration DNO run on UltraFeedback in terms of GPT-4o API calls and wall-clock time. By construction the pipeline makes about $2n$ API calls for $n$ prompts: $n$ calls to generate $\ygold$, and, because the three prompt partitions are disjoint and each prompt's six responses are rated in one batched query, a further $n$ rating calls over the three iterations, before any retries. For the 63{,}967 UltraFeedback prompts this estimate is roughly 128{,}000 calls; the count recorded in 2024, approximately 126{,}000, is slightly lower. Failed calls are retried up to three times.

Each iteration uses one third of the prompts, for which we generate five on-policy responses each with vLLM \citep{kwon2023vllm}; generation takes about three to four hours on four A100 GPUs. Rating all responses takes five to six hours of API time, which multithreading reduces to about two hours. Training takes six to eight hours on the same GPUs, depending on the number of large-margin pairs. Summed over three iterations with multithreaded rating, the listed stages take 33 to 42 hours; the end-to-end time recorded in 2024 was 35 to 45 hours.

\FloatBarrier
\section{Limitations}
\label{sec:limitations}

\textbf{The teacher is also the evaluation judge and an opponent:}
GPT-4o generates $\ygold$, rates every candidate, is one of the two AlpacaEval opponents and is the AlpacaEval judge. Between 23\% and 67\% of the selected pairs in each iteration have a GPT-4o response as the positive (Table~\ref{result:ablation:teacher-student-evolution}), the GPT-4o response is always the last candidate in the rating query, and LLM judges prefer outputs that resemble their own \citep{panickssery2024selfpref,wataoka2024selfpreference}. \citet{li2026preferenceleakage} measure an average preference-leakage score of about 24\% when judge and data generator are the same model, as in our main runs, and about 3--9\% when they share a model family. Those magnitudes come from other settings and are not a correction for ours, but they show that part of the measured gain may be GPT-4o-likeness as scored by GPT-4o rather than better responses. The open-weight-annotator runs of Section~\ref{result:larger-annotators} are a partial control only for the rating step, since the judge is unchanged: with Llama~3.1 70B Instruct as annotator the Llama student gains 8.4 LC points against GPT-4o, versus 13.6 with GPT-4o as teacher, and the difference cannot be separated into annotator quality and judge affinity. The comparison with the 70B model (Section~\ref{sec:results:larger}) is therefore a benchmark comparison under one protocol, not a ranking. A judge outside the GPT family, an independent benchmark, matched-length analysis, and a capability-regression check on standard benchmarks are the most important additions.

\textbf{Non-standard evaluation and single runs:}
Our evaluation protocol departs from the public leaderboard (Section~\ref{setup:general-setup}), so our numbers are not comparable to those in other papers, and the agreement of LC win rates with human rankings has not been established for it. Every cell is a single run, and between-run variability is unknown. A verbatim search finds 13 of the 805 AlpacaEval instructions (10 of them at least 40 characters long) in the processed UltraFeedback data; because a match may fall inside a response, about 1.6\% is an upper bound on exact prompt overlap, and near-duplicates were not searched. \citet{zheng2025nullmodels} show that a constant, adversarially crafted output reaches 86.5\% LC on AlpacaEval~2.0; this does not show that our checkpoints exploit the benchmark, but high absolute LC values, including our 70.18\% against GPT-4-0613, are evidence of preference under this judge rather than of quality in general.

\textbf{Hyperparameter selection:}
The learning rate was chosen after the first epoch or iteration of each experiment. If the selection criterion was the evaluation we report, the selection is optimistic; selecting on a separate split, with the reported evaluation kept out of the selection, would remove this optimism.

\textbf{Human preferences:}
No human-preference study is reported. Both the training signal and the evaluation are LLM judgments used as proxies for human preference, and the handling of intransitive preferences, the motivation for general-preference methods, was not tested.

\textbf{Baselines:}
SPIN is approximated by the teacher-always-positive rule rather than by the original recipe; SFT on UltraFeedback is a weak baseline for an instruct model; our SPPO and online DPO runs received only the learning-rate search and ended below the base model, so they say nothing about those methods; and no SimPO baseline was run.

\textbf{Response length:}
In the direct GPT-4o-annotated runs, response length grows with iterations, to 3{,}786 characters for Llama and 7{,}506 for Mistral (Section~\ref{sec:results}), and the rating step has no length control. We did not measure truncation for the Mistral runs or validate the length-control regression for these unusually long responses. Length growth alone does not demonstrate reward hacking, but it is the pattern reward hacking would produce; win rates at matched length would make the claim of more informative responses testable.

\textbf{A 2024 snapshot:}
The experiments were run in 2024 with the models and APIs of that time. GPT-4o is no longer a frontier judge, Mistral~7B Instruct~v0.1 is superseded, and neither concurrent Nash-style methods such as INPO \citep{zhang2025inpo} nor evaluator-trained open-weight judges \citep{wang2024selftaught} were compared against. We report the study because the procedural description and the design observations do not depend on the models being current; nothing has been re-run on newer models.

\textbf{Algorithmic approximations:}
Section~\ref{sec:ablation:errors} lists where DNO-Prct departs from the theory it approximates; in addition, none of our experiments evaluates the uniform mixture of iterates for which the Nash guarantee holds.

\section{Conclusion}
\label{sec:conclusion}
We re-implemented Direct Nash Optimization from its published description and studied it on two open-weight instruct models with proprietary and open-weight annotators. Under our evaluation protocol, three DNO iterations raise the length-controlled AlpacaEval~2.0 win rate against GPT-4o from 31.71\% to 45.32\% for Llama~3.1 8B Instruct and from 6.77\% to 18.44\% for Mistral~7B Instruct~v0.1, and the tuned 8B model scores above Llama~3.1 70B Instruct on all four reported win rates. The rating step does not require a proprietary model: same-family open-weight annotators retain 62\% and 82\% of the LC gain of the GPT-4o-annotated runs against GPT-4o, and against GPT-4-0613 the Llama run retains 45\% and the Mistral run matches the GPT-4o-annotated run. The ablations indicate which components matter: removing the gold responses or the look-back, or accepting only teacher-preferred pairs, scores 6 to 10 LC points below the full procedure, whereas the number of on-policy samples moves the LC win rate by at most 2.1 points and, in a separate sweep, the rating margin by at most 2.4 points. The defaults we would therefore keep are the teacher's gold responses, look-back, and rating teacher and student responses together so that a student response can be the positive, with response length measured explicitly. These findings rest on single runs evaluated by a GPT-4o judge that is also the teacher; an independent judge, repeated runs and the matched controls of Sections~\ref{sec:ablation:gold} and~\ref{sec:ablation:errors} are the natural next steps.

\section*{Acknowledgements}
A large language model was used to assist with editing this manuscript; the authors take full responsibility for its content.

\bibliographystyle{icml2024}
\bibliography{main}

\begin{thebibliography}{67}
\providecommand{\natexlab}[1]{#1}
\providecommand{\url}[1]{\texttt{#1}}
\expandafter\ifx\csname urlstyle\endcsname\relax
  \providecommand{\doi}[1]{doi: #1}\else
  \providecommand{\doi}{doi: \begingroup \urlstyle{rm}\Url}\fi

\bibitem[Amodei et~al.(2016)Amodei, Olah, Steinhardt, Christiano, Schulman, and
  Man{\'e}]{amodei2016concrete}
Amodei, D., Olah, C., Steinhardt, J., Christiano, P., Schulman, J., and
  Man{\'e}, D.
\newblock Concrete problems in {AI} safety.
\newblock \emph{arXiv preprint arXiv:1606.06565}, 2016.

\bibitem[Azar et~al.(2024)Azar, Rowland, Piot, Guo, Calandriello, Valko, and
  Munos]{azar2023ipo}
Azar, M.~G., Rowland, M., Piot, B., Guo, D., Calandriello, D., Valko, M., and
  Munos, R.
\newblock A general theoretical paradigm to understand learning from human
  preferences.
\newblock In \emph{Proceedings of the 27th International Conference on
  Artificial Intelligence and Statistics}, volume 238 of \emph{Proceedings of
  Machine Learning Research}, pp.\  4447--4455. PMLR, 2024.
\newblock arXiv:2310.12036.

\bibitem[Bai et~al.(2022)Bai, Jones, Ndousse, Askell, Chen, et~al.]{bai2022hh}
Bai, Y., Jones, A., Ndousse, K., Askell, A., Chen, A., et~al.
\newblock Training a helpful and harmless assistant with reinforcement learning
  from human feedback.
\newblock \emph{arXiv preprint arXiv:2204.05862}, 2022.

\bibitem[Brown et~al.(2020)Brown, Mann, Ryder, Subbiah, Kaplan,
  et~al.]{brown2020gpt3}
Brown, T.~B., Mann, B., Ryder, N., Subbiah, M., Kaplan, J., et~al.
\newblock Language models are few-shot learners.
\newblock In \emph{Advances in Neural Information Processing Systems},
  volume~33, pp.\  1877--1901, 2020.
\newblock arXiv:2005.14165.

\bibitem[Calandriello et~al.(2024)Calandriello, Guo, Munos, Rowland, Tang,
  Avila~Pires, Richemond, Le~Lan, Valko, Liu, Joshi, Zheng, and
  Piot]{calandriello2024online}
Calandriello, D., Guo, Z.~D., Munos, R., Rowland, M., Tang, Y., Avila~Pires,
  B., Richemond, P.~H., Le~Lan, C., Valko, M., Liu, T., Joshi, R., Zheng, Z.,
  and Piot, B.
\newblock Human alignment of large language models through online preference
  optimisation.
\newblock In \emph{Proceedings of the 41st International Conference on Machine
  Learning}, volume 235 of \emph{Proceedings of Machine Learning Research},
  pp.\  5409--5435. PMLR, 2024.
\newblock arXiv:2403.08635.

\bibitem[Chen et~al.(2025)Chen, Wei, Zhu, Feng, and Meng]{chen2025evaluators}
Chen, W.-L., Wei, Z., Zhu, X., Feng, S., and Meng, Y.
\newblock Do {LLM} evaluators prefer themselves for a reason?
\newblock \emph{arXiv preprint arXiv:2504.03846}, 2025.

\bibitem[Chen et~al.(2024)Chen, Deng, Yuan, Ji, and Gu]{chen2024spin}
Chen, Z., Deng, Y., Yuan, H., Ji, K., and Gu, Q.
\newblock Self-play fine-tuning converts weak language models to strong
  language models.
\newblock In \emph{Proceedings of the 41st International Conference on Machine
  Learning}, volume 235 of \emph{Proceedings of Machine Learning Research},
  pp.\  6621--6642. PMLR, 2024.
\newblock arXiv:2401.01335.

\bibitem[Cheng et~al.(2024)Cheng, Yang, Li, Dai, Hu, Cao, Du, and
  Li]{cheng2024apo}
Cheng, P., Yang, Y., Li, J., Dai, Y., Hu, T., Cao, P., Du, N., and Li, X.
\newblock Adversarial preference optimization: Enhancing your alignment via
  {RM}-{LLM} game.
\newblock In \emph{Findings of the Association for Computational Linguistics:
  ACL 2024}, pp.\  3705--3716, Bangkok, Thailand, 2024. Association for
  Computational Linguistics.
\newblock \doi{10.18653/v1/2024.findings-acl.221}.
\newblock arXiv:2311.08045.

\bibitem[Chiang et~al.(2024)Chiang, Zheng, Sheng, Angelopoulos, Li, Li, Zhang,
  Zhu, Jordan, Gonzalez, and Stoica]{chiang2024chatbotarena}
Chiang, W.-L., Zheng, L., Sheng, Y., Angelopoulos, A.~N., Li, T., Li, D.,
  Zhang, H., Zhu, B., Jordan, M., Gonzalez, J.~E., and Stoica, I.
\newblock {Chatbot Arena}: An open platform for evaluating {LLM}s by human
  preference.
\newblock In \emph{Proceedings of the 41st International Conference on Machine
  Learning}, volume 235 of \emph{Proceedings of Machine Learning Research},
  pp.\  8359--8388. PMLR, 2024.
\newblock arXiv:2403.04132.

\bibitem[Christiano et~al.(2017)Christiano, Leike, Brown, Martic, Legg, and
  Amodei]{christiano2017deep}
Christiano, P.~F., Leike, J., Brown, T.~B., Martic, M., Legg, S., and Amodei,
  D.
\newblock Deep reinforcement learning from human preferences.
\newblock In \emph{Advances in Neural Information Processing Systems},
  volume~30, 2017.
\newblock arXiv:1706.03741.

\bibitem[Cui et~al.(2024)Cui, Yuan, Ding, Yao, He, Zhu, Ni, Xie, Xie, Lin, Liu,
  and Sun]{cui2023ultrafeedback}
Cui, G., Yuan, L., Ding, N., Yao, G., He, B., Zhu, W., Ni, Y., Xie, G., Xie,
  R., Lin, Y., Liu, Z., and Sun, M.
\newblock {UltraFeedback}: Boosting language models with scaled {AI} feedback.
\newblock In \emph{Proceedings of the 41st International Conference on Machine
  Learning}, volume 235 of \emph{Proceedings of Machine Learning Research},
  pp.\  9722--9744. PMLR, 2024.
\newblock arXiv:2310.01377.

\bibitem[Dong et~al.(2024)Dong, Xiong, Pang, Wang, Zhao, Zhou, Jiang, Sahoo,
  Xiong, and Zhang]{dong2024rlhfworkflow}
Dong, H., Xiong, W., Pang, B., Wang, H., Zhao, H., Zhou, Y., Jiang, N., Sahoo,
  D., Xiong, C., and Zhang, T.
\newblock {RLHF} workflow: From reward modeling to online {RLHF}.
\newblock \emph{Transactions on Machine Learning Research}, 2024.
\newblock arXiv:2405.07863.

\bibitem[Dubois et~al.(2024)Dubois, Galambosi, Liang, and
  Hashimoto]{dubois2024lcalpacaeval}
Dubois, Y., Galambosi, B., Liang, P., and Hashimoto, T.~B.
\newblock Length-controlled {AlpacaEval}: A simple way to debias automatic
  evaluators.
\newblock \emph{arXiv preprint arXiv:2404.04475}, 2024.

\bibitem[Fishburn(1984)]{fishburn1984}
Fishburn, P.~C.
\newblock Probabilistic social choice based on simple voting comparisons.
\newblock \emph{The Review of Economic Studies}, 51\penalty0 (4):\penalty0
  683--692, 1984.
\newblock \doi{10.2307/2297786}.

\bibitem[Gao et~al.(2024)Gao, Chang, Zhan, Oertell, Swamy, Brantley, Joachims,
  Bagnell, Lee, and Sun]{gao2024rebel}
Gao, Z., Chang, J.~D., Zhan, W., Oertell, O., Swamy, G., Brantley, K.,
  Joachims, T., Bagnell, J.~A., Lee, J.~D., and Sun, W.
\newblock {REBEL}: Reinforcement learning via regressing relative rewards.
\newblock In \emph{Advances in Neural Information Processing Systems},
  volume~37, pp.\  52354--52400, 2024.
\newblock arXiv:2404.16767.

\bibitem[Grattafiori et~al.(2024)Grattafiori, Dubey, Jauhri, Pandey, Kadian,
  et~al.]{grattafiori2024llama3}
Grattafiori, A., Dubey, A., Jauhri, A., Pandey, A., Kadian, A., et~al.
\newblock The {L}lama 3 herd of models.
\newblock \emph{arXiv preprint arXiv:2407.21783}, 2024.

\bibitem[Guo et~al.(2024)Guo, Zhang, Liu, Liu, Khalman, Llinares, Rame,
  Mesnard, Zhao, Piot, Ferret, and Blondel]{guo2024oaif}
Guo, S., Zhang, B., Liu, T., Liu, T., Khalman, M., Llinares, F., Rame, A.,
  Mesnard, T., Zhao, Y., Piot, B., Ferret, J., and Blondel, M.
\newblock Direct language model alignment from online {AI} feedback.
\newblock \emph{arXiv preprint arXiv:2402.04792}, 2024.

\bibitem[Holtzman et~al.(2020)Holtzman, Buys, Du, Forbes, and
  Choi]{holtzman2020nucleus}
Holtzman, A., Buys, J., Du, L., Forbes, M., and Choi, Y.
\newblock The curious case of neural text degeneration.
\newblock In \emph{International Conference on Learning Representations}, 2020.
\newblock arXiv:1904.09751.

\bibitem[Hu et~al.(2022)Hu, Shen, Wallis, Allen-Zhu, Li, Wang, Wang, and
  Chen]{hu2022lora}
Hu, E.~J., Shen, Y., Wallis, P., Allen-Zhu, Z., Li, Y., Wang, S., Wang, L., and
  Chen, W.
\newblock {LoRA}: Low-rank adaptation of large language models.
\newblock In \emph{International Conference on Learning Representations}, 2022.
\newblock URL \url{https://openreview.net/forum?id=nZeVKeeFYf9}.

\bibitem[Ivison et~al.(2024)Ivison, Wang, Liu, Wu, Pyatkin, Lambert, Smith,
  Choi, and Hajishirzi]{ivison2024unpacking}
Ivison, H., Wang, Y., Liu, J., Wu, Z., Pyatkin, V., Lambert, N., Smith, N.~A.,
  Choi, Y., and Hajishirzi, H.
\newblock Unpacking {DPO} and {PPO}: Disentangling best practices for learning
  from preference feedback.
\newblock In \emph{Advances in Neural Information Processing Systems},
  volume~37, 2024.
\newblock arXiv:2406.09279.

\bibitem[Jiang et~al.(2023)Jiang, Sablayrolles, Mensch, Bamford, Chaplot,
  de~las Casas, Bressand, Lengyel, Lample, Saulnier, Lavaud, Lachaux, Stock,
  Le~Scao, Lavril, Wang, Lacroix, and El~Sayed]{jiang2023mistral7b}
Jiang, A.~Q., Sablayrolles, A., Mensch, A., Bamford, C., Chaplot, D.~S., de~las
  Casas, D., Bressand, F., Lengyel, G., Lample, G., Saulnier, L., Lavaud,
  L.~R., Lachaux, M.-A., Stock, P., Le~Scao, T., Lavril, T., Wang, T., Lacroix,
  T., and El~Sayed, W.
\newblock Mistral 7{B}.
\newblock \emph{arXiv preprint arXiv:2310.06825}, 2023.

\bibitem[Jiang et~al.(2024)Jiang, Sablayrolles, Roux, Mensch, Savary,
  et~al.]{jiang2024mixtral}
Jiang, A.~Q., Sablayrolles, A., Roux, A., Mensch, A., Savary, B., et~al.
\newblock Mixtral of experts.
\newblock \emph{arXiv preprint arXiv:2401.04088}, 2024.

\bibitem[Kramer(1973)]{kramer1973}
Kramer, G.~H.
\newblock On a class of equilibrium conditions for majority rule.
\newblock \emph{Econometrica}, 41\penalty0 (2):\penalty0 285--297, 1973.
\newblock \doi{10.2307/1913490}.

\bibitem[Kreweras(1965)]{kreweras1965}
Kreweras, G.
\newblock Aggregation of preference orderings.
\newblock In \emph{Mathematics and Social Sciences {I}: Proceedings of the
  Seminars of Menthon-Saint-Bernard, France (1--27 July 1960) and of
  G{\"o}sing, Austria (3--27 July 1962)}, pp.\  73--79. Mouton, Paris, 1965.

\bibitem[Kwon et~al.(2023)Kwon, Li, Zhuang, Sheng, Zheng, Yu, Gonzalez, Zhang,
  and Stoica]{kwon2023vllm}
Kwon, W., Li, Z., Zhuang, S., Sheng, Y., Zheng, L., Yu, C.~H., Gonzalez, J.~E.,
  Zhang, H., and Stoica, I.
\newblock Efficient memory management for large language model serving with
  {PagedAttention}.
\newblock In \emph{Proceedings of the 29th ACM Symposium on Operating Systems
  Principles (SOSP)}, 2023.
\newblock arXiv:2309.06180.

\bibitem[Lanchantin et~al.(2025)Lanchantin, Chen, Lan, Li, Saha, Wang, Xu, Yu,
  Yuan, Weston, Sukhbaatar, and Kulikov]{lanchantin2025bridging}
Lanchantin, J., Chen, A., Lan, J., Li, X., Saha, S., Wang, T., Xu, J., Yu, P.,
  Yuan, W., Weston, J.~E., Sukhbaatar, S., and Kulikov, I.
\newblock Bridging offline and online reinforcement learning for {LLMs}.
\newblock \emph{arXiv preprint arXiv:2506.21495}, 2025.

\bibitem[Li et~al.(2026)Li, Sun, Huang, Zhong, Jiang, Han, Zhang, Wang, and
  Liu]{li2026preferenceleakage}
Li, D., Sun, R., Huang, Y., Zhong, M., Jiang, B., Han, J., Zhang, X., Wang, W.,
  and Liu, H.
\newblock Preference leakage: A contamination problem in {LLM}-as-a-judge.
\newblock In \emph{The Fourteenth International Conference on Learning
  Representations}, 2026.
\newblock arXiv:2502.01534.

\bibitem[Li et~al.(2023)Li, Zhang, Dubois, Taori, Gulrajani, Guestrin, Liang,
  and Hashimoto]{li2023alpacaeval}
Li, X., Zhang, T., Dubois, Y., Taori, R., Gulrajani, I., Guestrin, C., Liang,
  P., and Hashimoto, T.~B.
\newblock {AlpacaEval}: An automatic evaluator of instruction-following models.
\newblock \url{https://github.com/tatsu-lab/alpaca_eval}, 2023.

\bibitem[Liu et~al.(2024)Liu, Zhou, Liu, Bu, Yang, Zhong, and
  Ouyang]{liu2024iterative}
Liu, J., Zhou, Z., Liu, J., Bu, X., Yang, C., Zhong, H.-S., and Ouyang, W.
\newblock Iterative length-regularized direct preference optimization: A case
  study on improving {7B} language models to {GPT-4} level.
\newblock \emph{arXiv preprint arXiv:2406.11817}, 2024.

\bibitem[Liu et~al.(2025)Liu, Long, Shi, Su, and Xiao]{liu2025condorcet}
Liu, K., Long, Q., Shi, Z., Su, W.~J., and Xiao, J.
\newblock Statistical impossibility and possibility of aligning {LLMs} with
  human preferences: From {Condorcet} paradox to {Nash} equilibrium.
\newblock \emph{arXiv preprint arXiv:2503.10990}, 2025.

\bibitem[Meng et~al.(2024)Meng, Xia, and Chen]{meng2024simpo}
Meng, Y., Xia, M., and Chen, D.
\newblock {SimPO}: Simple preference optimization with a reference-free reward.
\newblock In \emph{Advances in Neural Information Processing Systems},
  volume~37, 2024.
\newblock arXiv:2405.14734.

\bibitem[{Meta AI}(2024)]{meta2024llamarecipes}
{Meta AI}.
\newblock {Llama Recipes} (now {Llama Cookbook}): Scripts for fine-tuning {Meta
  Llama} models.
\newblock \url{https://github.com/meta-llama/llama-recipes}, 2024.
\newblock Accessed 2026\mbox{-}10\mbox{-}14.

\bibitem[{Mistral AI}(2024)]{mistral2024large}
{Mistral AI}.
\newblock Large enough.
\newblock \url{https://mistral.ai/news/mistral-large-2407/}, 2024.
\newblock Announcement of Mistral Large 2 (Mistral-Large-Instruct-2407, 123B
  parameters). Accessed 2026\mbox{-}10\mbox{-}14.

\bibitem[Munos et~al.(2024)Munos, Valko, Calandriello, Azar, Rowland, Guo,
  Tang, Geist, Mesnard, Michi, Selvi, Girgin, Momchev, Bachem, Mankowitz,
  Precup, and Piot]{munos2023nlhf}
Munos, R., Valko, M., Calandriello, D., Azar, M.~G., Rowland, M., Guo, Z.~D.,
  Tang, Y., Geist, M., Mesnard, T., Michi, A., Selvi, M., Girgin, S., Momchev,
  N., Bachem, O., Mankowitz, D.~J., Precup, D., and Piot, B.
\newblock {N}ash learning from human feedback.
\newblock In \emph{Proceedings of the 41st International Conference on Machine
  Learning}, volume 235 of \emph{Proceedings of Machine Learning Research},
  pp.\  36743--36768. PMLR, 2024.
\newblock arXiv:2312.00886.

\bibitem[{OpenAI}(2023)]{openai2023gpt4}
{OpenAI}.
\newblock {GPT-4} technical report.
\newblock \emph{arXiv preprint arXiv:2303.08774}, 2023.

\bibitem[{OpenAI}(2024)]{openai2024gpt4o}
{OpenAI}.
\newblock {GPT-4o} system card.
\newblock \emph{arXiv preprint arXiv:2410.21276}, 2024.

\bibitem[{OpenBMB}(2023)]{openbmb2023ultrafeedbackdata}
{OpenBMB}.
\newblock {UltraFeedback} dataset.
\newblock \url{https://huggingface.co/datasets/openbmb/UltraFeedback}, 2023.
\newblock Accessed 2026\mbox{-}10\mbox{-}14.

\bibitem[Ouyang et~al.(2022)Ouyang, Wu, Jiang, Almeida, Wainwright, Mishkin,
  Zhang, Agarwal, Slama, Ray, Schulman, Hilton, Kelton, Miller, Simens, Askell,
  Welinder, Christiano, Leike, and Lowe]{ouyang2022instructgpt}
Ouyang, L., Wu, J., Jiang, X., Almeida, D., Wainwright, C.~L., Mishkin, P.,
  Zhang, C., Agarwal, S., Slama, K., Ray, A., Schulman, J., Hilton, J., Kelton,
  F., Miller, L., Simens, M., Askell, A., Welinder, P., Christiano, P., Leike,
  J., and Lowe, R.
\newblock Training language models to follow instructions with human feedback.
\newblock In \emph{Advances in Neural Information Processing Systems},
  volume~35, pp.\  27730--27744, 2022.
\newblock arXiv:2203.02155.

\bibitem[Panickssery et~al.(2024)Panickssery, Bowman, and
  Feng]{panickssery2024selfpref}
Panickssery, A., Bowman, S.~R., and Feng, S.
\newblock {LLM} evaluators recognize and favor their own generations.
\newblock In \emph{Advances in Neural Information Processing Systems},
  volume~37, pp.\  68772--68802, 2024.
\newblock arXiv:2404.13076.

\bibitem[Park et~al.(2024)Park, Rafailov, Ermon, and Finn]{park2024length}
Park, R., Rafailov, R., Ermon, S., and Finn, C.
\newblock Disentangling length from quality in direct preference optimization.
\newblock In \emph{Findings of the Association for Computational Linguistics:
  ACL 2024}, pp.\  4998--5017, Bangkok, Thailand, 2024. Association for
  Computational Linguistics.
\newblock \doi{10.18653/v1/2024.findings-acl.297}.
\newblock arXiv:2403.19159.

\bibitem[Radford et~al.(2018)Radford, Narasimhan, Salimans, and
  Sutskever]{radford2018gpt}
Radford, A., Narasimhan, K., Salimans, T., and Sutskever, I.
\newblock Improving language understanding by generative pre-training.
\newblock Technical report, OpenAI, 2018.
\newblock URL
  \url{https://cdn.openai.com/research-covers/language-unsupervised/language_understanding_paper.pdf}.

\bibitem[Radford et~al.(2019)Radford, Wu, Child, Luan, Amodei, and
  Sutskever]{radford2019gpt2}
Radford, A., Wu, J., Child, R., Luan, D., Amodei, D., and Sutskever, I.
\newblock Language models are unsupervised multitask learners.
\newblock Technical report, OpenAI, 2019.
\newblock URL
  \url{https://cdn.openai.com/better-language-models/language_models_are_unsupervised_multitask_learners.pdf}.

\bibitem[Rae et~al.(2021)Rae, Borgeaud, Cai, Millican, Hoffmann,
  et~al.]{rae2021gopher}
Rae, J.~W., Borgeaud, S., Cai, T., Millican, K., Hoffmann, J., et~al.
\newblock Scaling language models: Methods, analysis \& insights from training
  {G}opher.
\newblock \emph{arXiv preprint arXiv:2112.11446}, 2021.

\bibitem[Rafailov et~al.(2023)Rafailov, Sharma, Mitchell, Ermon, Manning, and
  Finn]{rafailov2023dpo}
Rafailov, R., Sharma, A., Mitchell, E., Ermon, S., Manning, C.~D., and Finn, C.
\newblock Direct preference optimization: Your language model is secretly a
  reward model.
\newblock In \emph{Advances in Neural Information Processing Systems},
  volume~36, pp.\  53728--53741, 2023.
\newblock arXiv:2305.18290.

\bibitem[Ross et~al.(2011)Ross, Gordon, and Bagnell]{ross2011dagger}
Ross, S., Gordon, G., and Bagnell, D.
\newblock A reduction of imitation learning and structured prediction to
  no-regret online learning.
\newblock In \emph{Proceedings of the Fourteenth International Conference on
  Artificial Intelligence and Statistics}, volume~15 of \emph{Proceedings of
  Machine Learning Research}, pp.\  627--635. PMLR, 2011.
\newblock URL \url{https://proceedings.mlr.press/v15/ross11a.html}.

\bibitem[Rosset et~al.(2024)Rosset, Cheng, Mitra, Santacroce, Awadallah, and
  Xie]{rosset2024dno}
Rosset, C., Cheng, C.-A., Mitra, A., Santacroce, M., Awadallah, A., and Xie, T.
\newblock Direct {N}ash optimization: Teaching language models to self-improve
  with general preferences.
\newblock \emph{arXiv preprint arXiv:2404.03715}, 2024.

\bibitem[Sharma et~al.(2024)Sharma, Keh, Mitchell, Finn, Arora, and
  Kollar]{sharma2024critical}
Sharma, A., Keh, S., Mitchell, E., Finn, C., Arora, K., and Kollar, T.
\newblock A critical evaluation of {AI} feedback for aligning large language
  models.
\newblock In \emph{Advances in Neural Information Processing Systems},
  volume~37, 2024.
\newblock arXiv:2402.12366.

\bibitem[Simpson(1969)]{simpson1969}
Simpson, P.~B.
\newblock On defining areas of voter choice: {P}rofessor {T}ullock on stable
  voting.
\newblock \emph{The Quarterly Journal of Economics}, 83\penalty0 (3):\penalty0
  478--490, 1969.
\newblock \doi{10.2307/1880533}.

\bibitem[Stiennon et~al.(2020)Stiennon, Ouyang, Wu, Ziegler, Lowe, Voss,
  Radford, Amodei, and Christiano]{stiennon2020summarize}
Stiennon, N., Ouyang, L., Wu, J., Ziegler, D.~M., Lowe, R., Voss, C., Radford,
  A., Amodei, D., and Christiano, P.
\newblock Learning to summarize with human feedback.
\newblock In \emph{Advances in Neural Information Processing Systems},
  volume~33, pp.\  3008--3021, 2020.
\newblock arXiv:2009.01325.

\bibitem[Swamy et~al.(2024)Swamy, Dann, Kidambi, Wu, and Agarwal]{swamy2024spo}
Swamy, G., Dann, C., Kidambi, R., Wu, Z.~S., and Agarwal, A.
\newblock A minimaximalist approach to reinforcement learning from human
  feedback.
\newblock In \emph{Proceedings of the 41st International Conference on Machine
  Learning}, volume 235 of \emph{Proceedings of Machine Learning Research},
  pp.\  47345--47377. PMLR, 2024.
\newblock arXiv:2401.04056.

\bibitem[Tang et~al.(2026)Tang, Yoon, Son, Yuan, Gu, and
  Bogunovic]{tang2025rspo}
Tang, X., Yoon, S., Son, S., Yuan, H., Gu, Q., and Bogunovic, I.
\newblock {RSPO}: Regularized self-play alignment of large language models.
\newblock In \emph{Proceedings of the 43rd International Conference on Machine
  Learning}, 2026.

\bibitem[Tang et~al.(2024)Tang, Guo, Zheng, Calandriello, Cao, Tarassov, Munos,
  {\'A}vila~Pires, Valko, Cheng, and Dabney]{tang2024online}
Tang, Y., Guo, D.~Z., Zheng, Z., Calandriello, D., Cao, Y., Tarassov, E.,
  Munos, R., {\'A}vila~Pires, B., Valko, M., Cheng, Y., and Dabney, W.
\newblock Understanding the performance gap between online and offline
  alignment algorithms.
\newblock \emph{arXiv preprint arXiv:2405.08448}, 2024.

\bibitem[Touvron et~al.(2023)Touvron, Martin, Stone, Albert, Almahairi,
  et~al.]{touvron2023llama2}
Touvron, H., Martin, L., Stone, K., Albert, P., Almahairi, A., et~al.
\newblock {L}lama 2: Open foundation and fine-tuned chat models.
\newblock \emph{arXiv preprint arXiv:2307.09288}, 2023.

\bibitem[Tunstall et~al.(2023)Tunstall, Beeching, Lambert, Rajani, Rasul,
  Belkada, Huang, von Werra, Fourrier, Habib, Sarrazin, Sanseviero, Rush, and
  Wolf]{tunstall2023zephyr}
Tunstall, L., Beeching, E., Lambert, N., Rajani, N., Rasul, K., Belkada, Y.,
  Huang, S., von Werra, L., Fourrier, C., Habib, N., Sarrazin, N., Sanseviero,
  O., Rush, A.~M., and Wolf, T.
\newblock Zephyr: Direct distillation of {LM} alignment.
\newblock \emph{arXiv preprint arXiv:2310.16944}, 2023.

\bibitem[von Werra et~al.(2020)von Werra, Belkada, Tunstall, Beeching, Thrush,
  Lambert, Huang, Rasul, and Gallou{\'e}dec]{vonwerra2020trl}
von Werra, L., Belkada, Y., Tunstall, L., Beeching, E., Thrush, T., Lambert,
  N., Huang, S., Rasul, K., and Gallou{\'e}dec, Q.
\newblock {TRL}: Transformers reinforcement learning.
\newblock \url{https://github.com/huggingface/trl}, 2020.

\bibitem[Wang et~al.(2025)Wang, Ma, Chen, Meng, Han, Xiao, Zhang, Huo, Su, and
  Yang]{wang2025mpo}
Wang, M., Ma, C., Chen, Q., Meng, L., Han, Y., Xiao, J., Zhang, Z., Huo, J.,
  Su, W.~J., and Yang, Y.
\newblock Magnetic preference optimization: Achieving last-iterate convergence
  for language model alignment.
\newblock In \emph{The Thirteenth International Conference on Learning
  Representations}, 2025.
\newblock arXiv:2410.16714.

\bibitem[Wang et~al.(2024)Wang, Kulikov, Golovneva, Yu, Yuan, Dwivedi-Yu, Pang,
  Fazel-Zarandi, Weston, and Li]{wang2024selftaught}
Wang, T., Kulikov, I., Golovneva, O., Yu, P., Yuan, W., Dwivedi-Yu, J., Pang,
  R.~Y., Fazel-Zarandi, M., Weston, J., and Li, X.
\newblock Self-taught evaluators.
\newblock \emph{arXiv preprint arXiv:2408.02666}, 2024.

\bibitem[Wataoka et~al.(2024)Wataoka, Takahashi, and
  Ri]{wataoka2024selfpreference}
Wataoka, K., Takahashi, T., and Ri, R.
\newblock Self-preference bias in {LLM}-as-a-judge.
\newblock \emph{arXiv preprint arXiv:2410.21819}, 2024.

\bibitem[Wu et~al.(2025)Wu, Sun, Yuan, Ji, Yang, and Gu]{wu2024sppo}
Wu, Y., Sun, Z., Yuan, H., Ji, K., Yang, Y., and Gu, Q.
\newblock Self-play preference optimization for language model alignment.
\newblock In \emph{The Thirteenth International Conference on Learning
  Representations}, 2025.
\newblock arXiv:2405.00675.

\bibitem[Xiong et~al.(2024)Xiong, Dong, Ye, Wang, Zhong, Ji, Jiang, and
  Zhang]{xiong2024iterative}
Xiong, W., Dong, H., Ye, C., Wang, Z., Zhong, H., Ji, H., Jiang, N., and Zhang,
  T.
\newblock Iterative preference learning from human feedback: Bridging theory
  and practice for {RLHF} under {KL}-constraint.
\newblock In \emph{Proceedings of the 41st International Conference on Machine
  Learning}, volume 235 of \emph{Proceedings of Machine Learning Research},
  pp.\  54715--54754. PMLR, 2024.
\newblock arXiv:2312.11456.

\bibitem[Xu et~al.(2023)Xu, Lee, Sukhbaatar, and Weston]{xu2023cringe}
Xu, J., Lee, A., Sukhbaatar, S., and Weston, J.
\newblock Some things are more {CRINGE} than others: Iterative preference
  optimization with the pairwise cringe loss.
\newblock \emph{arXiv preprint arXiv:2312.16682}, 2023.

\bibitem[Yuan et~al.(2024)Yuan, Pang, Cho, Li, Sukhbaatar, Xu, and
  Weston]{yuan2024selfrewarding}
Yuan, W., Pang, R.~Y., Cho, K., Li, X., Sukhbaatar, S., Xu, J., and Weston, J.
\newblock Self-rewarding language models.
\newblock In \emph{Proceedings of the 41st International Conference on Machine
  Learning}, volume 235 of \emph{Proceedings of Machine Learning Research},
  pp.\  57905--57923. PMLR, 2024.
\newblock arXiv:2401.10020.

\bibitem[Zhang et~al.(2025{\natexlab{a}})Zhang, Yu, Ge, Song, Zeng, Mi, Jiang,
  and Yu]{zhang2025onpo}
Zhang, Y., Yu, D., Ge, T., Song, L., Zeng, Z., Mi, H., Jiang, N., and Yu, D.
\newblock Improving {LLM} general preference alignment via optimistic online
  mirror descent.
\newblock In \emph{Advances in Neural Information Processing Systems},
  2025{\natexlab{a}}.

\bibitem[Zhang et~al.(2025{\natexlab{b}})Zhang, Yu, Peng, Song, Tian, Huo,
  Jiang, Mi, and Yu]{zhang2025inpo}
Zhang, Y., Yu, D., Peng, B., Song, L., Tian, Y., Huo, M., Jiang, N., Mi, H.,
  and Yu, D.
\newblock Iterative {N}ash policy optimization: Aligning {LLM}s with general
  preferences via no-regret learning.
\newblock In \emph{The Thirteenth International Conference on Learning
  Representations}, 2025{\natexlab{b}}.
\newblock arXiv:2407.00617.

\bibitem[Zhang et~al.(2025{\natexlab{c}})Zhang, Zhang, Wu, Xu, and
  Gu]{zhang2025gpm}
Zhang, Y., Zhang, G., Wu, Y., Xu, K., and Gu, Q.
\newblock Beyond {Bradley-Terry} models: A general preference model for
  language model alignment.
\newblock In \emph{Proceedings of the 42nd International Conference on Machine
  Learning}, 2025{\natexlab{c}}.

\bibitem[Zheng et~al.(2023)Zheng, Chiang, Sheng, Zhuang, Wu, Zhuang, Lin, Li,
  Li, Xing, Zhang, Gonzalez, and Stoica]{zheng2023judge}
Zheng, L., Chiang, W.-L., Sheng, Y., Zhuang, S., Wu, Z., Zhuang, Y., Lin, Z.,
  Li, Z., Li, D., Xing, E.~P., Zhang, H., Gonzalez, J.~E., and Stoica, I.
\newblock Judging {LLM}-as-a-judge with {MT-Bench} and {Chatbot Arena}.
\newblock In \emph{Advances in Neural Information Processing Systems},
  volume~36, pp.\  46595--46623, 2023.
\newblock Datasets and Benchmarks Track. arXiv:2306.05685.

\bibitem[Zheng et~al.(2025)Zheng, Pang, Du, Liu, Jiang, and
  Lin]{zheng2025nullmodels}
Zheng, X., Pang, T., Du, C., Liu, Q., Jiang, J., and Lin, M.
\newblock Cheating automatic {LLM} benchmarks: Null models achieve high win
  rates.
\newblock In \emph{The Thirteenth International Conference on Learning
  Representations}, 2025.
\newblock arXiv:2410.07137.

\end{thebibliography}

\clearpage
\onecolumn
\appendix

\section{Rating Prompt}
\label{app:prompt}
The annotator receives the prompt of Figure~\ref{fig:prompt}, reproduced verbatim from our code. The user prompt replaces \texttt{\{instructions\}}, and the candidate responses replace \texttt{\{candidate\_answers\}}, the $K$ on-policy samples first (five unless stated) and $\ygold$ last, indexed from zero; the doubled braces are escapes of Python's string formatting and reach the annotator as single braces. The rubric is additive over six criteria, one point each, so scores range from 0 to 6, and the annotator is asked to return only a Python dictionary with one score per candidate. Our prompt follows \citet{rosset2024dno} with modifications so that it transfers to annotators outside the GPT family.

\begin{figure}[ht]
\lstset{basicstyle=\ttfamily\fontsize{6.2}{7.5}\selectfont,breaklines=false,columns=fixed,keepspaces=true,frame=single,framesep=5pt,rulecolor=\color{black!40},showstringspaces=false,aboveskip=0pt,belowskip=0pt}
\begin{lstlisting}
Review the user's question and the corresponding response using the additive 6-point scoring system described below. Points
are accumulated based on the satisfaction of each criterion:
1. Give 1 point if the response is formatted as an answer, e.g. does not repeat itself over and over, is not a continuation of
the question or other random question/answer pairs or examples, etc.
2. Add 1 more point if the response is relevant and provides some information related to the user's inquiry, even if it is
incomplete or contains some irrelevant content.
3. Add a third point if the response addresses a substantial portion of the user's question without rambling, but does not
completely resolve the query or provide a direct answer.
4. Award a fourth point if the response answers nearly all the elements of the user's question correctly, in a useful and
direct way, and is largely devoid of errors, extraneous content, or excess wordiness.
5. Grant a fifth point if the response is entirely correct (without any errors or made-up facts), and directly answers all
aspects of the user's question in a helpful, well-organized, focused, and succinct style.
6. Bestow a sixth point for a response that is entirely correct *AND* concise, impeccably tailored to the user's question,
without any unnecessary information or wordiness, reflecting expert knowledge, amounting to a superbly high-quality,
focused, and insightful answer.
### User: {instructions}
### Candidate Responses:
{candidate_answers}

### Instructions: After examining the user's instruction and the response:
1. Score EACH answer using the format: "Score: <totalpoints>"
Remember to assess from the AI Assistant perspective, utilizing web search knowledge as necessary. To evaluate the response
in alignment with this additive scoring model, we'll systematically attribute points based on the outlined criteria. Remember to
score EACH candidate answer separately. Format your score only as a dictionary with no extraneous output
e.g. {{"id0_score": 3, "id1_score": 5}} with "idX" corresponding to the id of the X'th answer (zero-indexed).
2. Give only a python dict without containing any other words. For example, you should output the following:
{{"id0_score": 3, "id1_score": 5, "id2_score": 6}}

### Output:
\end{lstlisting}
\caption{The rating prompt used in all DNO runs, verbatim from our code.}
\label{fig:prompt}
\end{figure}

\section{Hyperparameters}
\label{app:hparams}
Table~\ref{tab:hparams} lists the training, generation and evaluation settings.

\begin{table}[ht]
\centering
\small
\setlength{\tabcolsep}{4pt}
\caption{Training, generation and evaluation hyperparameters.}
\label{tab:hparams}
\begin{tabular}{>{\raggedright\arraybackslash}p{0.30\columnwidth} >{\raggedright\arraybackslash}p{0.62\columnwidth}}
\toprule
Hyperparameter & Value \\
\midrule
Fine-tuning & LoRA: rank 16, $\alpha$ 32, dropout 0.05; adapters merged after each DNO iteration \\
Learning rate (SFT) & $5\times10^{-5}$ \\
Learning rate (DPO, DNO) & $5\times10^{-6}$ (most DNO runs) \\
Learning-rate search & $\{5\times10^{-5}, 5\times10^{-6}\}$, first epoch / iteration \\
Optimizer; scheduler & AdamW; constant after 100 warm-up steps; weight decay 0 \\
Effective batch size & $2$ per device $\times$ $4$ GPUs $\times$ $4$ accumulation steps $= 32$ \\
GPUs & 4 $\times$ NVIDIA A100-SXM 80\,GB \\
Epochs (SFT / DPO) & 20 (checkpoints at 3, 10) / 3 \\
DNO iterations $\times$ epochs per iteration & $3 \times 1$ \\
DPO loss; coefficient $\beta$ & sigmoid; 0.1 \\
Reference policy & $\pi_t$: the iteration's starting checkpoint with adapters disabled \\
Max.\ sequence / prompt length & 4{,}096 / 3{,}072 tokens (training) \\
Seed & 2024 \\
On-policy samples per prompt & 5 \\
Sampling temperature / top-$p$ & 0.7 / 0.95 \\
Rating query & all candidates in one query, fixed order, $\ygold$ last \\
Positive threshold & rating $\ge 5$ on the 0--6 scale \\
Rating margin ($\pTS$, $\pSS$ / $\pST$) & 2 / 1 points \\
Data look-back & at most 30\% of the pairs of $t-1$, 15\% of $t-2$, \ldots \\
Step cap (margin ablation) & 2{,}000 training steps per iteration \\
Prompt partitions & 3 disjoint splits of UltraFeedback \\
Evaluation decoding & vLLM, chat template, temperature 0.001, at most 4{,}096 new tokens \\
AlpacaEval & v0.6.5; references by GPT-4o and GPT-4-0613 (temperature 0); judge GPT-4o, \texttt{alpaca\_eval\_gpt4} ranking template \\
Software & TRL 0.8.6, transformers 4.43.3, PEFT 0.10.0, accelerate 0.32.1, datasets 2.16.1; vLLM \\
\bottomrule
\end{tabular}
\end{table}

\section{Pair-Construction Procedure}
\label{app:pairs}
Algorithm~\ref{alg:pairs} states one DNO-Prct iteration with the thresholds of Section~\ref{setup:dno-data}. Ratings are integers from 0 to 6; a response is a positive if its rating is at least five. Candidates are kept in a fixed order, the $K$ student samples first and the gold response last, and the test $y^{-} = \ygold_x$ is on the index, not on the text. $D^{\mathrm{new}}_t$ denotes the pairs built from iteration $t$'s own samples, before look-back.

\begin{algorithm}[ht]
\caption{One DNO-Prct iteration (pair construction and update)}
\label{alg:pairs}
\begin{algorithmic}[1]
\REQUIRE prompt partition $X_t$; current policy $\pi_t$; gold responses $\{\ygold_x\}$; annotator $A$; samples per prompt $K$ ($K = 5$ unless stated); margin $m = 2$; relaxed margin $m' = 1$; look-back fractions $(0.30, 0.15, \ldots)$
\STATE $D^{\mathrm{new}}_t \leftarrow \emptyset$
\FOR{each prompt $x \in X_t$}
  \STATE sample $y_1, \ldots, y_K \sim \pi_t(\cdot \mid x)$ with temperature $0.7$, top-$p$ $0.95$; drop samples containing degenerate character sequences
  \STATE $C \leftarrow (y_1, \ldots, y_K, \ygold_x)$; query $A$ once with $x$ and $C$; skip $x$ unless the reply gives one score $s(y)$ per $y \in C$
  \STATE $P \leftarrow \{y \in C : s(y) \ge 5\}$
  \FOR{each $y^{+} \in P$ and each $y^{-} \in C \setminus \{y^{+}\}$}
    \STATE $\mu \leftarrow m'$ \textbf{if} $y^{+}$ is a student sample \textbf{and} $y^{-} = \ygold_x$ \textbf{else} $m$
    \IF{$s(y^{+}) - s(y^{-}) \ge \mu$}
      \STATE $D^{\mathrm{new}}_t \leftarrow D^{\mathrm{new}}_t \cup \{(x, y^{+}, y^{-})\}$
    \ENDIF
  \ENDFOR
\ENDFOR
\STATE $D_t \leftarrow D^{\mathrm{new}}_t$
\FOR{$k = 1, 2, \ldots, t-1$}
  \STATE add to $D_t$ at most a fraction $0.30 \cdot 2^{-(k-1)}$ of $D_{t-k}$, the training set of iteration $t-k$ (which includes its own look-back pairs), unchanged
\ENDFOR
\STATE train LoRA adapters on $\pi_t$ with the DPO loss ($\beta = 0.1$) on $D_t$ for one epoch (2{,}000-step cap in the margin sweep), with reference policy $\pi_t$; merge them to obtain $\pi_{t+1}$
\end{algorithmic}
\end{algorithm}

\end{document}